%% file: main.tex
\documentclass[conference]{IEEEtran}
\IEEEoverridecommandlockouts
\usepackage{cite}
\usepackage{amsmath,amssymb,amsfonts}
\usepackage{algorithm}
\usepackage{algorithmic}
\usepackage{graphicx}
\usepackage{textcomp}
\usepackage{xcolor}
\usepackage{microtype}
\usepackage{graphicx}
\usepackage{subfigure}
\usepackage{booktabs} 

\usepackage{amsmath} 
\usepackage{amsthm}
\usepackage{amssymb}
\usepackage{makecell}
\usepackage{multicol}
\usepackage{multirow}
\usepackage{graphicx}
\usepackage{dblfloatfix}
\usepackage{float}
\usepackage{enumitem}
\usepackage{arydshln}
\usepackage{hyperref}
\usepackage{authblk}

\theoremstyle{plain}
\newtheorem{thm}{Theorem}
\newtheorem{lem}{Lemma}
\newtheorem{prop}{Proposition}

\newcommand{\sg}{\textsc {SUGAR} }

\newcommand\norm[1]{{\left\Vert#1\right\Vert}_{\infty}}
\newcommand\fnorm[1]{{\left\Vert#1\right\Vert}_{F}}
\def\cl{\mathcal{L}}
\def\ie{\textit{i.e.}}
\def\eg{\textit{e.g.}}
\def\st{\textit{s.t.}}

\usepackage{amsmath}
\usepackage{amssymb}
\usepackage{mathtools}
\usepackage{amsthm}

\usepackage[capitalize,noabbrev]{cleveref}

\theoremstyle{plain}

\theoremstyle{definition}

\theoremstyle{remark}

\usepackage[textsize=tiny]{todonotes}
\def\BibTeX{{\rm B\kern-.05em{\sc i\kern-.025em b}\kern-.08em
    T\kern-.1667em\lower.7ex\hbox{E}\kern-.125emX}}

\begin{document}

\title{SUGAR: Efficient Subgraph-level Training via Resource-aware Graph Partitioning
}

\author{Zihui Xue}
\author{Yuedong Yang}
\author{Mengtian Yang}
\author{Radu Marculescu}
\affil[]{The University of Texas at Austin}

\maketitle

\begin{abstract}
Graph Neural Networks (GNNs) have demonstrated a great potential in a variety of graph-based applications, such as recommender systems, drug discovery, and object recognition. Nevertheless, resource-efficient GNN learning is a rarely explored topic despite its many benefits for edge computing and Internet of Things (IoT) applications. To improve this state of affairs, this work proposes efficient \underline{su}b\underline{g}raph-level tr\underline{a}ining via \underline{r}esource-aware graph partitioning (SUGAR). SUGAR first partitions the initial graph into a set of disjoint subgraphs and then performs local training at the subgraph-level. We provide a theoretical analysis and conduct extensive experiments on five graph benchmarks to verify its efficacy in practice. Our results show that SUGAR can achieve up to 33× runtime speedup and 3.8× memory reduction on large-scale graphs. We believe SUGAR opens a new research direction towards developing GNN methods that are resource-efficient, hence suitable for IoT deployment.

\end{abstract}

\input{1.intro}
\input{2.related}
\input{3.method_new}
\input{4.exp}
\input{5.conclusion}

\bibliography{main}
\bibliographystyle{plain}
\input{6.appendix}

\end{document}

%% file: 1.intro.tex
\section{Introduction}
\label{sec.intro}
Graphs are non-Euclidean data structures that can model complex relationships among a set of interacting objects, for instance, social networks, knowledge graphs, or biological networks. Given the huge success of deep neural networks for Euclidean data (\eg, images, text and audio), there is an increasing interest in developing deep learning approaches for graphs too. Graph Neural Networks (GNNs) generalize the convolution operation to the non-Euclidean domain \cite{wu2020comprehensive}; they demonstrate a great potential for various graph-based applications, such as node classification \cite{gcn}, link prediction \cite{zhang2018link} and recommender systems \cite{fan2019graph}. 

The rapid development of smart devices and IoT applications has spawned a great interest in many edge AI applications. Training models locally becomes a growing trend as this can help avoid data transmission to the cloud, reduce communication latency, and better preserve privacy \cite{flsurvey}. For instance, in a graph-based recommender system, user data can be quite sensitive and hence it's better to store it locally \cite{wu2021fedgnn}. This brings about the need for \textit{resource-efficient graph learning}.

While there is much discussion about locally training Convolutional Neural Networks (CNNs) \cite{Bagherinezhad_2017_CVPR}, efficient on-device training for GNNs is rarely explored. Different from CNNs, where popular models such as ResNet \cite{resnet} are deep and have a large parameter space, mainstream GNN models are shallow and more lightweight. However, the major bottleneck of GNN training comes from the nodes dependencies in the input graph. Consequently, graph convolution suffers from a high computational cost, as the representation of a node in the current layer needs to be computed recursively by the representations of all neighbors in its previous layer. Moreover, storing the intermediate features for all nodes requires much memory space, especially when the graph size grows. For instance, for the \textit{ogbn-products} graph in our experiments (Table \ref{tab.dataset}), full-batch training requires a GPU with 33GB of memory \cite{ogb}. Thus scaling GNN training to large-scale graphs remains a big challenge. The problem is more severe for a resource-constrained scenario like IoT, where GNN training is heavily constrained by the computation, memory, and communication costs. 


Various approaches have been proposed to alleviate the computation and memory burden of GNNs. For instance, sampling-based approaches aim at reducing the neighborhood size via layer sampling \cite{graphsage, fastgcn, chen2017stochastic}, clustering based sampling \cite{clustergcn} and graph sampling \cite{graphsaint} techniques; these prior works approach this problem purely from an algorithmic angle. A few recent works \cite{gist, distdgl} investigate the topic of distributed multi-GPU training of GNNs and achieve good parallel efficiency and memory scalability while using large GPU clusters.

A common limitation of all these approaches is that they \textit{do not} take the real hardware constraints into consideration. For mobile devices with limited memory budgets, the input graph can be too large to fit entirely in the main memory. In addition, the communication overhead among real IoT devices is significantly larger than when using GPU clusters, rendering distributed training approaches not readily applicable to such scenarios. This calls for a new approach for \textit{resource-efficient GNN learning}, which is precisely the focus of our paper.
 

In this work, we propose a novel approach that trains GNNs efficiently with multiple devices in a resource-limited scenario. To this end, we (1) design a graph partitioning method that accounts for resource constraints and graph topology; (2) train a set of local GNNs at the \textit{subgraph-level} for computation, memory and communication savings. Our contributions are as follows:
\begin{itemize}[itemsep=2pt,topsep=4pt,parsep=2pt]
\item We formulate the problem of training GNNs with multiple resource-constrained devices. Although our formulation targets various mobile and edge devices (\eg, mobile phones, Raspberry Pi), it is also applicable to powerful machines equipped with GPUs.
\item We propose \textsc{SUGAR}, a GNN training framework that aims at improving training scalability. We provide complexity analysis, error bound and convergence analysis of the proposed estimator. 
\item We show that \sg achieves the best runtime and memory usage (with similar accuracy) when compared against state-of-the-art GNN approaches on five large-scale datasets and across multiple hardware platforms, ranging from edge devices (\ie, Raspberry Pi, Jetson Nano) to a desktop equipped with powerful GPUs.
\item We illustrate the flexibility of \sg by integrating it with both full-batch and mini-batch algorithms such as GraphSAGE \cite{graphsage} and GraphSAINT \cite{graphsaint}. Experimental results demonstrate that \sg can achieve up to 33× runtime speedup on \textit{ogbn-arxiv} and 3.8× memory reduction on \textit{Reddit}. On the \textit{ogbn-products} graph with over 2 million nodes and 61 million edges, \sg achieves 1.62× speedup over GraphSAGE and 1.83× memory reduction over GraphSAINT with a better test accuracy ($\sim$0.7\%). 
\end{itemize}

The remainder of the paper is organized as follows. In Section 2, we discuss prior work. In Section 3, we formulate the problem and describe our proposed training framework \textsc{SUGAR}. Experimental results are presented in Section 4. Finally, Section 5 concludes the paper.

%% file: 2.related.tex
\section{Related Work}
\label{sec.related}
The relevant prior work comes from three directions as discussed next.
\subsection{Graph Neural Networks}
Modern GNNs adopt a neighborhood aggregation scheme to learn representations for individual nodes or the entire graph. Graph Convolution Network (GCN) \cite{gcn} is a pioneering work that generalizes the use of regular convolutions to graphs. GraphSAGE \cite{graphsage} provides an inductive graph representation learning framework. To improve the representation ability of GNNs, Graph Attention Networks (GAT) \cite{gat} introduce self-attention to the graph convolution operation. Apart from pursuing higher accuracy, a few GNN architecture improvements \cite{sign, wu2019simplifying} have been made towards higher training efficiency.


\subsection{GNN training algorithms}

Current GNN training algorithms can be categorized into full-batch training and mini-batch training. 
 
\textbf{Full-batch training} was first proposed for GCNs \cite{gcn}; the gradient is calculated based on the global graph and updated once per epoch. Despite being fast, full-batch gradient descent is generally infeasible for large-scale graphs due to excessively large memory requirements and slow convergence.

\textbf{Mini-batch training} was first proposed in GraphSAGE \cite{graphsage}; the gradient update is based on a proportion of nodes in the graph and updated a few times during each training epoch. Mini-batch training leads to memory efficiency at the cost of increased computation. Since the neighborhood aggregation scheme involves recursive calculation of a node's neighbors layer by layer, time complexity becomes exponential with respect to the number of GNN layers; this is known as the \textit{neighborhood expansion problem}.  


Following the idea of neighbor sampling, FastGCN \cite{fastgcn} further proposes the importance node sampling to reduce variance. The work of \cite{chen2017stochastic} proposes a control variate based algorithm that allows a smaller neighbor sample size. 

A few recent works propose alternative ways to construct mini-batches instead of layer-wise sampling. For instance, ClusterGCN \cite{clustergcn} first partitions the training graph into clusters and then randomly groups clusters together as a batch. GraphSAINT \cite{graphsaint} builds mini-batches by sampling the training graph and ensures a fixed number of nodes in all layers.

\subsection{Graph Sparsification}
Recent works have also investigated graph sparsification (\ie, pruning edges of the training graph) for GNN learning.  In many real-world applications, graphs exhibit complex topology patterns. Some edges may be erroneous or task-irrelevant, and thus aggregating this information weakens the generalizability of GNNs \cite{luo2021learning}. As shown by \cite{rong2019dropedge} and \cite{zhao2019pairnorm}, edges of the input graph may be pruned without loss of accuracy.

Two recent works introduce computation efficiency into the problem. More precisely, SGCN \cite{li2020sgcn} proposes a neural network that prunes edges of the input graph; they show that using sparsified graphs as the new input for GNNs brings computational benefits. UGS \cite{chen2021unified} presents a graph lottery ticket type of approach; they sparsify the input graph, as well as model weights during training to save inference computation. 

%% file: 3.method_new.tex
\section{Our Proposed Method}\label{sec.method}
\subsection{Problem Formulation} \label{sec.form}
Given a graph $\mathcal{G} = (\mathcal V,\mathcal E)$, where $\mathcal V$ is the node set and $\mathcal E$ represents the set of edges. Let $N=|\mathcal V|$ denote the number of nodes and $A \in R^{N\times N}$ be the adjacency matrix of $\mathcal{G}$. Every node $i$ is characterized by a $F$-dimensional feature vector $x_i \in R^{F}$. We use $X \in R^{N \times F}$ to represent the feature matrix of all nodes in $\mathcal G$. 

Consider a node-level prediction problem with the following objective: 
\begin{equation}
\begin{aligned}
    \min_W \cl &= \frac{1}{N}\sum_{i=1}^N f(y_{i}, z_{i}) \\
    z_i &= g(x_i; W)
\end{aligned}
\label{eq.orig_formulation}
\end{equation}
where $f$ is the objective function (\eg, cross entropy for node classification), $y_i$ and $z_i$ denotes the true label and prediction of node $i$, respectively. $g(\cdot)$ denotes a graph neural network parameterized by $W$ that generates node-level predictions.





Suppose there are $K$ devices available for training, and let $\mathcal B_{MEM}^k$ denote the memory budget of device $k$. Motivated by the notorious inefficiency that centralized graph learning suffers from, we aim at \textit{distributing the training process} to improve the training scalability. The key is to assign $N$ nodes of graph $\mathcal G$ to $K$ devices, and then do local training on \textit{each} device. We formulate it as two subproblems below. 

First, we define a graph partitioning strategy $\mathcal{P}: \mathcal V \rightarrow (\mathcal V_1, \mathcal V_2, \cdots, \mathcal V_K)$ that divides the node set $\mathcal V$ into $K$ subsets such that: 
\begin{equation}
	\begin{split}
	 	\cup_k \mathcal V_k  = \mathcal V,\ \ H(\mathcal{SG}_k) < \mathcal B_{MEM}^k, \ \  \forall k \in [K]
	\end{split}
	\label{eq.constraint}
\end{equation}
where $[K]=\{1,...,K\}$, $\mathcal{SG}_k$ denotes the subgraph induced by node set $\mathcal V_i$, $H$ is a static function that maps a given subgraph $\mathcal{SG}_i$ to the device memory requirements for training. For maximum generality, here we do not require $\mathcal V_i \cap \mathcal V_j = \varnothing$. In other words, a node $i$ can be assigned to more than one hardware device, and let $\mathcal P_i$ denote the set of hardware devices where node $i$ is assigned to. 


Next, we adopt subgraph-level training, \ie, for device $k$, we maintain a local GNN model, denoted by $ W^{\langle k \rangle}$ that takes the subgraph $\mathcal{SG}_k$ as its input graph. Let $W=\frac{1}{K}\sum_{k=1}^K  W^{\langle k \rangle}$, thus the objective can be reformulated as:
\begin{equation}
\begin{aligned}
    \min_{W} \cl &= \frac{1}{N}\sum_{i=1}^N f(y_{i}, z_{i}) \\
    z_i &= \frac{1}{|\mathcal P_i|} \sum_{k\in \mathcal P_i}g(x_i;  W^{\langle k \rangle})
\end{aligned}
\label{eq.dist_formulation}
\end{equation}

\begin{figure*}[thp]
\centering
\includegraphics[width=0.85\textwidth]{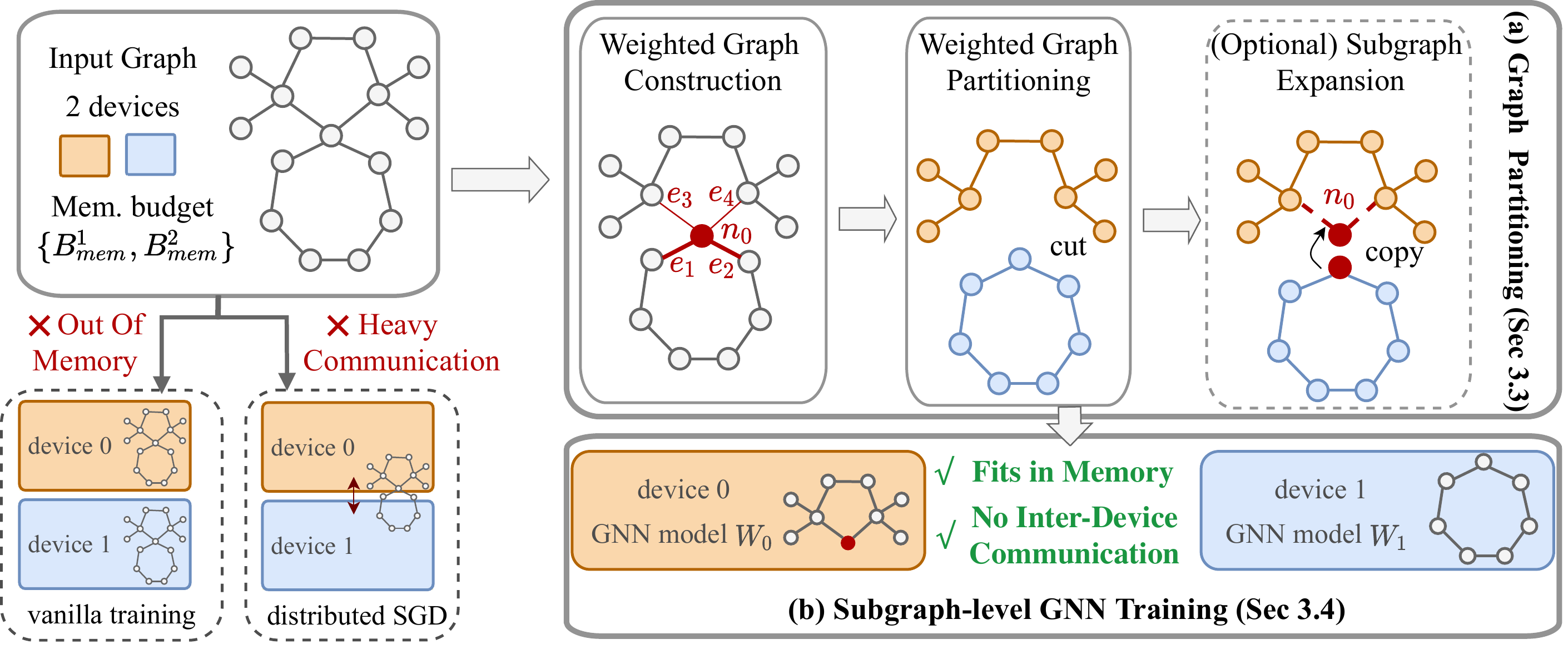} 
\caption{While vanilla training is likely to run out of memory when the graph size is large and distributed stochastic gradient descent (SGD) requires heavy intermediate communication among devices, \sg provides a solution that is memory efficient and requires no inter-device communication. The proposed \sg consists of two stages: (a) graph partitioning and (b) subgraph-level GNN training. Graph partitioning involves three steps: (1) transform the input graph $\mathcal{G}$ to a weighted graph $\mathcal{G}^w$; (2) apply METIS to the weighted graph $\mathcal{G}^w$, where edges with large weights are more likely to be preserved; (3) (optional) expand the node set of the obtained subgraph according to memory budgets.}
\label{fig.framework}
\end{figure*}

Based on the formulation above, we propose \textsc{SUGAR}, a distributed training framework that: (1) partitions the input graph subject to resource constraints; (2) adopts local subgraph-level training. Figure \ref{fig.framework} provides a simple illustration of \textsc{SUGAR} for a two-device system. We describe our design choices in detail in the following sections.


\subsection{Theoretical Basis}\label{subsec.theory}
Recall that we define a graph partitioning strategy $\mathcal P$ that divides $N$ nodes into $K$ node sets $(\mathcal V_1, \mathcal V_2, \cdots, \mathcal V_K)$. Taking $K$ subgraphs induced by the node sets into consideration, a graph partitioning strategy $\mathcal P$ can be viewed as a way to produce a sparser adjacency matrix $A_{SG}$, from the original matrix $A$. 
$A_{SG}$ is a block-diagonal matrix of $A$, \ie, 
\begin{equation}
  A_{SG} = \begin{bmatrix}
		A_{\mathcal V_1} & \cdots & 0 & \cdots & 0 \\
		\vdots & \ddots &  &  & \vdots \\
		0 & & A_{\mathcal V_k} & & 0\\
		\vdots & & & \ddots & \vdots \\ 
		0 & \cdots & 0 & \cdots & A_{\mathcal V_K}
	\end{bmatrix}  
\end{equation}
where $A_{\mathcal V_k}$ denotes the adjacency matrix of subgraph $k$. 

We show below that adopting $A_{SG}$ for training offers the benefits of high computational efficiency and low memory requirements. Moreover, we provide the error bound and convergence analysis of this approximation for a graph convolutional network (GCN) \cite{gcn}. 

\textbf{Complexity Analysis.} The propagation rule for the $l$-th layer GCN is:
\begin{equation}
    Z^{(l+1)} = A^{norm}H^{(l)}W^{(l)}, H^{(l+1)} = \sigma(Z^{(l+1)})
\end{equation}

where $\sigma$ represents an activation function, $A^{norm}$ denotes the normalized version of $A$, \ie, $A^{norm}=\hat D^{-1/2}\hat A \hat D^{1/2}, \hat A = A + I_N, \hat D_{ii}=\sum_j\hat A_{ij}$ and $I_N$ is an $N$-dimensional identity matrix. $H^{(l)}$ and $H^{(l+1)}$ denotes the input and output feature matrices in layer $l$, respectively. $Z^{(l)}$ is the node feature matrix before the activation function in layer $l$ and $Z^{(L)}$ denotes final node predictions (\ie, output of the GCN). $W^{(l)}\in R^{F_l\times F_{l+1}}$ represents the weight matrix of layer $l$, where $F_l$ and $F_{l+1}$ is the input and output feature dimension, respectively. Therefore, for the $l$-th layer GCN, the \textit{training time complexity} is $\mathcal O(|\mathcal E|F_l + NF_l F_{l+1})$ and \textit{memory complexity} is $\mathcal O(NF_{l+1}+F_l F_{l+1})$. We make two observations here: (a) Real-world graphs are usually sparse and $\frac{|\mathcal E|}{N}$ is generally smaller than feature number $F_{l+1}$. Thus, the second term dominates the time complexity; (b) For large-scale graphs, the number of nodes $N$ is much greater than the number of features. Consequently, $\mathcal O(NF_{l+1})$ dominates the memory complexity. 
It is easy to verify that the number of nodes $N$ imposes a computation hurdle on training. Partitioning the input graph into $K$ subgraphs reduces the number of nodes $N$ to $N_k=|\mathcal{V}_k|$ for every local model. Since $N_k $ is about $1/K$ of $N$, the proposed approach is expected to achieve up to $K$ times speedup, and as little as $1/K$ of the original memory requirements.

\textbf{Error Bound Analysis.}
Let our proposed estimator be SG. The $l$-th layer propagation rule of a GCN with the SG estimator is:
\begin{equation}
    Z_{SG}^{(l+1)} = A_{SG}^{norm}H_{SG}^{(l)}W^{(l)}, H_{SG}^{(l+1)} = \sigma(Z_{SG}^{(l+1)})
\end{equation}
where $Z_{SG}^{(l+1)}$ and $H_{SG}^{(l+1)}$ denote the node representations produced by the SG estimator in layer $l+1$ before and after activation, respectively. 

Assume that we run graph partitioning for $M$ times to obtain a sample average of $A_{SG}^{norm}$ before training. Let $\epsilon=\norm{A_{SG}^{norm}-A^{norm}}$ denote the error in approximating $A^{norm}$ with $A_{SG}^{norm}$. For simplicity, we will omit the superscript $norm$ from now on. 

The following lemma states that the error of node predictions given by the SG estimator is bounded.

\begin{lem}
For a multi-layer GCN with fixed weights, assume that: (1) $\sigma(\cdot)$ is $\rho$-Lipschitz and $\sigma(0) = 0$, (2) input matrices $A$, $X$ and model weights $\{W^{(l)}\}_{l=1}^L$ are all bounded, 
then there exists $C$ such that $\norm{Z_{SG}^{(l)}-Z^{(l)}} \leq C\epsilon, \forall l \in [L]$ and $\norm{H_{SG}^{(l)}-H^{(l)}} \leq C\epsilon, \forall l \in [L-1]$.
\label{lem.eb_activation}
\end{lem}
The proof of Lemma \ref{lem.eb_activation} is provided in Appendix \ref{apsec.proof1}. Lemma \ref{lem.eb_activation} motivates us to design a graph partitioning method that generates small $\epsilon$ so that the output of the SG estimator is close to the exact value. This will be discussed in detail in the next subsection.

\textbf{Convergence Analysis.} 
Let $W_t$ denote the model parameters at training epoch $t$ and $W_*$ denote the optimal model weights. $\nabla \cl(W) = \frac{1}{N}\sum_{i=1}^N \frac{\partial f(y_i, z_i^{(L)})}{\partial W}$ and $\nabla \cl_{SG}(W) =\frac{1}{N}\sum_{i=1}^N \frac{\partial f(y_i, z_{SG, i}^{(L)})}{\partial W}$ represent the gradients of the exact GCN and SG estimator with respect to model weights $W$, respectively.

Theorem \ref{thm.converge} states that with high probability gradient descent training with the approximated gradients of the SG estimator (\ie, $\nabla \cl_{SG}(W)$) converges to a local minimum.
\begin{thm}
Assume that: (1) the loss function $\mathcal L(W)$ is $\rho$-smooth, (2) the gradients of the loss $\nabla \cl(W)$ and $\nabla \cl_{SG}(W)$ are bounded for any choice of $W$, (3) the gradient of the objective function $\frac{\partial f(y, z)}{\partial z}$ is $\rho$-Lipschitz and bounded, (4) the activation function $\sigma (\cdot)$ is $\rho$-Lipschitz, $\sigma(0) = 0$ and its gradient is bounded,

then there exists $C>0$, \st, $\forall M, T$, for a sufficiently small $\delta$, if we run graph partitioning for $M$ times and run gradient descent for $R \leq T$ epochs (where $R$ is chosen uniformly from $[T]$, the model update rule is $W_{t+1} = W_t - \gamma \nabla \mathcal L_{SG}(W_t)$, and step size $\gamma = \frac{1}{\rho\sqrt T}$), we have:
\begin{equation*}
\begin{aligned}
 &P(\mathbb E_{R} \fnorm{\nabla \cl(W_R)}^2 \leq \delta) \geq \\
 &1-2 \exp \{-2M (\frac{\delta}{2C} - \frac{2\rho [\cl(W_1) - \cl(W_*)] + C - \delta}{2C(\sqrt T -1)})^2 \}
\end{aligned}
\end{equation*}
\label{thm.converge}
\end{thm}
With $M$ and $T$ increasing, the right-hand-side of the inequality becomes larger. This implies that there is a higher probability for the loss to converge to a local minimum. The full proof is provided in Appendix \ref{apsec.proof1}.
\subsection{Graph Partitioning}
From Lemma \ref{lem.eb_activation}, we conclude that a graph partitioning method that yields a smaller $|A_{SG}-A|$ leads to a smaller error in node predictions. Therefore, we aim at minimizing the difference between $A_{SG}$ and $A$. In other words, the objective of graph partitioning should be to \textit{minimize the number of edges of the incident nodes} that belong to different subsets. As such, this is identical to the goal of various existing graph partitioning methods, making such approaches good candidates to use with our framework. We choose METIS \cite{metis} due to its efficiency in handling large-scale graphs. However, the traditional graph partitioning algorithms are \textit{not} intended for modern GNNs and the learning component of the problem is missing. Consequently, we present a modified version of METIS that is suited to our problem and relies on two new ideas discussed next.


\textbf{a) Weighted Graph Construction}. We build a weighted graph $\mathcal G^w$ from the input graph $\mathcal G$. The \textit{weight} of an edge $e_{uv}$ is defined based on the degree of its two incident nodes: 
\begin{equation}
 \begin{split}
	& weight(e_{uv}) = d_{max} + 1 - deg(u) - deg(v) \\
	& d_{max} = max\{deg(u)+deg(v), \forall e_{uv} \in \mathcal E \}
\end{split}
\label{eq.weighted_graph}
\end{equation}

Let $A^w$ denote the adjacency matrix of the weighted graph $\mathcal G^w$, where element $a_{ij}^w$ is the edge weight $weight(e_{ij})$; $a_{ij}^w$ is 0 if there is no edge connecting nodes $i$ and $j$. 


The key intuition behind our first idea lies in the neighborhood aggregation scheme of GNNs. Consider two nodes $u$ and $v$, where $u$ is a hub node connected to many other nodes, while $v$ has only one neighbor. As GNNs propagate by aggregating the neighborhood information of nodes, removing the only edge of node $v$ may possibly lead to wrong predictions. On the other hand, pruning an edge of $u$ is more acceptable since there are many neighbors contributing to its prediction. Consider the graph in Figure \ref{fig.framework} as an example. Cutting the edges $e_1\cup e_2$ and $e_3\cup e_4$ are both feasible solutions for METIS. However, considering the fact that nodes connected to $e_1$ and $e_2$ have less topology information, our proposed method will preserve them and cut edges $e_3\cup e_4$ instead; this can lead to a better learning performance.

As can be concluded from this small example, edges connected to small-degree nodes are critical to our problem and should be preserved. Conversely, edges connected to high-degree nodes may be intentionally ignored. This explains our weights definition strategy. Consequently, we incorporate the above observation into our partitioning objective and apply METIS to the pre-processed graph $\mathcal G^w$.

\textbf{b) Subgraph Expansion}. After obtaining the partitions with our modified METIS, we propose the second idea, \ie, expand the subgraph based on available hardware resources. Although METIS only provides partitioning results where the node sets do not overlap, our general formulation in Section \ref{sec.form} allows nodes to belong to multiple partitions. This brings great flexibility to our approach to adjust the node number for each device according to its memory budget. 

Suppose the available memory of device $k$ is larger than the actual requirement of training a GNN on subgraph $k$ (\ie, $H(\mathcal{SG}_k) < B_{MEM}^k$), then we may choose to expand the node set $\mathcal{V}_k$ by adding the one-hop neighbors of nodes that do not belong to $\mathcal{V}_k$. As illustrated in Figure \ref{fig.framework} (a), we can expand the node set of the subgraph on device 0 (marked in light brown) to include node $n_{0}$ as well. While expanding the subgraph is likely to yield higher accuracy, training time and memory requirement will also increase. Therefore, this is an optional step, only if the hardware resources allow it.
%

\subsection{Subgraph-level Local Training}
From the original formulation in Equation \ref{eq.dist_formulation}, if $|\mathcal P_i| > 1$, \ie, a node $i$ is assigned to multiple devices, calculating its loss and backpropagation can involve heavy communication among devices. To address this problem, we provide the following result to decouple the training of $K$ local GNN models from each other. 

\begin{prop}
If $f(y, z)$ is convex with respect to $z$, then the upper bound of $\cl$ in Equation \ref{eq.dist_formulation} is given by:
\begin{equation}
\begin{aligned}
\cl &\leq \frac{1}{K} \sum_{k=1}^K  \sum_{i\in \mathcal V_k} \frac{1}{|\mathcal P_i|} f(y_i, z_i) \\
 &z_i = g(x_i, W^{\langle k \rangle}) \\
\end{aligned}
\label{eq.loss-ub}
\end{equation}
\label{prop.ub}
\end{prop}
The proof is provided in Appendix \ref{apsec.proof2}.

Proposition \ref{prop.ub} allows us to shift the perspective from `node-level' to `device-level'. We adopt the upper bound of $\cl$ in Equation \ref{eq.loss-ub} as the new training objective. Now, the local model updates involving node $i$ \textit{do not} depend on other models (\ie, $\{W^{\langle k \rangle}\}_{k \in \mathcal P_i}$) any more. Optimizing the new objective naturally reduces the upper bound of the original one and avoids significant communication costs, thus leading to high training efficiency. 

Furthermore, motivated by deployment challenges in real IoT applications, where communication among devices is generally not guaranteed, we propose to reduce inter-device communication down to zero in our framework. In particular, we maintain $K$ distinct (local) models instead of a single (global) model by keeping the local model updates within each device. The objective of our proposed subgraph-level local GNN training can be summarized as follows:
\begin{equation}
\begin{aligned}
    \min_{W^{\langle k \rangle}} \cl_{k}&= \sum_{i\in \mathcal V_k} \frac{1}{|\mathcal P_i|} f(y_i, z_i), \ \forall k \in [K]\\
    z_i &= g(x_i, W^{\langle k \rangle})
    \end{aligned}
\end{equation}
In training round $t$, every device performs local updates as:
\begin{equation}
    W^{\langle k \rangle}_{t+1} \gets  W^{\langle k \rangle}_t - \gamma \nabla_{W^{\langle k \rangle}} \cl_{k},\ \forall k \in [K]
\end{equation}
where $\cl_{k}$ denotes the training objective of device $k$ and $\gamma$ is the learning rate (\ie, step size). By decoupling training dependency among devices, we propose a feasible solution to train GNNs in resource-limited scenarios, where typical distributed GNN approaches are not applicable.

\subsection{Putting it all together}
\begin{algorithm}[!htb]
   \caption{\sg}
   \label{alg.sgtrain}
   \hspace*{0.02in} {\bf Input:} 
   graph $\mathcal{G} = (\mathcal V,\mathcal E)$; node feature matrix $X$; available device number $K$; device memory budget $\{B_{MEM}^k\}_{k=1}^K$; total training epochs $T$.
\begin{algorithmic}[1]
   \STATE Construct $\mathcal G^w$ from $\mathcal G$ according to Equation \ref{eq.weighted_graph}
   \STATE Partition $\mathcal G^w$ into $K$ subgraphs $\{\mathcal{SG}_i\}_1^K$ 
   \STATE (Optional) Expand $\mathcal{SG}_i$ if $H(\mathcal{SG}_i) < B_{MEM}^i$
   \FOR{each device $k=\{1,2,\cdots,K\}$ in parallel}
   \STATE Initialize GNN model weight $W^{\langle k \rangle}_{1}$
   \FOR{epoch $t=1,2,\cdots,T$}
   		\STATE $ W^{\langle k \rangle}_{t+1} \gets  W^{\langle k \rangle}_t - \gamma \nabla_{W^{\langle k \rangle}} \cl_{k}$ 
   		\ENDFOR
   \ENDFOR
\end{algorithmic}
\end{algorithm}

To sum up, the \sg algorithm consists of two stages: (a) graph partitioning (lines 1-3) and (b) subgraph-level GNN training (lines 4-9). Specifically, the graph partitioning involves three steps: (1) construct a weighted graph $\mathcal G^w$ from $\mathcal G$ to account for the influence of node degrees in learning (line 1). (2) Apply METIS to the weighted graph $\mathcal G^w$ to obtain partitioning results (line 2). (3) According to the memory budget, expand the subgraph to cover the one-hop neighbors for better performance (line 3). Then, we train $K$ local models in parallel without requiring training-time communication among devices (lines 4-9). The proposed subgraph-level training with multiple devices achieves high training efficiency, low memory requirements and zero communication costs.




%% file: 4.exp.tex
\section{Experiments}\label{sec.exp}

\subsection{Experimental Setup}

We perform a thorough evaluation of \sg on five node classification benchmarks, \ie, \textit{Flickr}, \textit{Reddit}, \textit{ogbn-arxiv}, \textit{ogbn-proteins} and \textit{ogbn-products} \cite{snapnets, ogb}. Dataset statistics are summarized in Table \ref{tab.dataset}. More details of the datasets are in Appendix \ref{apsec.setup}.

\begin{table}[htb]
\caption{Dataset Statistics. K and M denote 1,000 and 1,000,000, respectively. `AvgDeg.' represents the average node degree. `ACC' denotes accuracy.}
\label{tab.dataset}
\vskip 0.1in
\begin{center}
\begin{small}
\begin{tabular}{lccccc}
\toprule
 Dataset & \textit{Flickr} & \textit{Reddit} & \makecell[c]{\textit{ogbn-} \\ \textit{arxiv}} & \makecell[l]{\textit{ogbn-} \\ \textit{proteins}} & \makecell[c]{\textit{ogbn-} \\ \textit{products}} \\
\midrule
\#Nodes                     & 89.3K  & 233K & 169K & 133K  & 2,449K \\
\#Edges                     & 0.90M & 11.6M & 1.17M  & 39.6M  & 61.9M \\
AvgDeg.  & 10 & 50 & 13.77  & 597  & 50.5 \\
\#Tasks                      & 1 & 1 & 1 & 112 & 1 \\
\#Classes                   & 7 & 41 & 40 & 2 & 47 \\
Metric                      & ACC  & ACC & ACC & \makecell[c]{ROC-\\AUC} & ACC \\
\bottomrule
\end{tabular}
\end{small}
\end{center}
\vskip -0.15in
\end{table}


We include the following GNN architectures and training algorithms for comparison: 
(1) GCN \cite{gcn},
(2) GraphSAGE \cite{graphsage}: mini-batch GraphSAGE are denoted by GraphSAGE-mb, 
(3) GAT \cite{gat},
(4) SIGN \cite{sign},
(5) ClusterGCN \cite{clustergcn},
(6) GraphSAINT \cite{graphsaint}: the random node, random edge, and random walk based samplers are denoted by GraphSAINT-N, GraphSAINT-E, GraphSAINT-RW, respectively. Descriptions about these GNN baselines are in Appendix \ref{apsec.setup}.


\sg is implemented with PyTorch \cite{pytorch} and DGL \cite{dgl}. For all the baseline methods, we use the parameters reported in their github pages or the original paper. We report accuracy results averaged over 5 runs for \textit{ogbn-proteins} and 10 runs for the other datasets. 

For completeness, we run our experiments across multiple hardware platforms. We select five different devices with various computing and memory capabilities, namely, (1) Raspberry Pi 3B, (2) NVIDIA Jetson Nano, (3) Android phone with Snapdragon 845 processor, (4) laptop with Intel i5-8279U CPU, and (5) desktop with AMD Threadripper 3970X CPU and two NVIDIA RTX 3090 GPUs.


\subsection{Results}
\subsubsection{Evaluations on GPUs}

First, we provide evaluation of \sg on a two-GPU system. Table \ref{tab.arxiv_gpu} and Table \ref{tab.reddit_gpu} report the average training time per epoch, maximum GPU memory usage and accuracy on \textit{ogbn-arxiv} and \textit{Reddit}. We base \sg on full-batch GCN and GraphSAGE for these two datasets, respectively. As shown in these tables, when compared with full-batch methods (\ie, GCN and GAT for \textit{ogbn-arxiv}; GraphSAGE for \textit{Reddit}), \sg is much more memory efficient, as it reduces the peak memory by $1.7\times$ for \textit{ogbn-arxiv} and $3.8\times$ for \textit{Reddit} data. When compared against mini-batch methods (\ie, mini-batch GraphSAGE, ClusterGCN, GraphSAINT and SIGN), the runtime of \sg is significantly smaller. This demonstrates the great benefits of our proposed subgraph-level training. Indeed, by restricting the neighborhood search size, \sg effectively alleviates the neighborhood expansion problem. In addition, it achieves very competitive test accuracies. 

\begin{table}[thb]
\caption{Runtime, memory \& accuracy results on \textit{ogbn-arxiv}. `Avg. Time' is the training time per epoch averaged over 100 epochs and `Max Mem' denotes peak allocated memory on GPU.}
\label{tab.arxiv_gpu}
\vskip 0.1in
\begin{center}
\begin{small}
\begin{tabular}{lcccc}
\toprule
  & \makecell[c]{Avg. \\ Time \\ \lbrack{\upshape ms}\rbrack} & \makecell[c]{\sg \\ Speedup} & \makecell[c]{Max \\ Mem \\ \lbrack GB\rbrack}  & \makecell[c]{Test \\ Acc. \\ \lbrack\%\rbrack } \\
\midrule
GCN  & 26.9  & 1.68$\times$ & 1.60 & 72.37 \scriptsize{$\pm$ 0.10}  \\
GAT  & 207.8 & 12.99$\times$ & 5.41 & 72.95 \scriptsize{$\pm$ 0.14} \\
GraphSAGE & 534.7 & 33.42$\times$ & 0.95 &  71.98 \scriptsize{$\pm$ 0.17}\\
SIGN & 291.6 & 18.23$\times$ & 0.94 &  71.79 \scriptsize{$\pm$ 0.08} \\
\hdashline
\textbf{\sg} & 16.0 &  & 0.92 & 72.22 \scriptsize{$\pm$ 0.14} \\
\bottomrule
\end{tabular}
\end{small}
\end{center}
\vskip -0.15in
\end{table}

\begin{table}[thb]
\caption{Runtime, memory \& accuracy results on \textit{Reddit}.}
\label{tab.reddit_gpu}
\vskip 0.1in
\begin{center}
\begin{small}
\begin{tabular}{lcccc}
\toprule
  & \makecell[c]{Avg. \\ Time \\ \lbrack {\upshape ms}\rbrack} & \makecell[c]{\sg \\ Speedup} & \makecell[c]{Max \\Mem \\ \lbrack GB\rbrack}  & \makecell[c]{Test\\ Acc. \\ \lbrack\%\rbrack} \\
\midrule
GraphSAGE & 110.6 & 1.87$\times$ & 5.70  & 96.39 \scriptsize{$\pm$ 0.03} \\
GraphSAGE-mb      & 316.5 & 5.36$\times$ & 2.33 & 95.08 \scriptsize{$\pm$ 0.05} \\
ClusterGCN    & 414.4 & 7.01$\times$ & 1.83 & 96.34 \scriptsize{$\pm$ 0.01} \\
GraphSAINT-N       & 341.8 & 5.78$\times$ & 1.29  & 96.17 \scriptsize{$\pm$ 0.06} \\
GraphSAINT-E      & 299.8 & 5.07$\times$ & 1.22  & 96.15 \scriptsize{$\pm$ 0.06} \\
GraphSAINT-RW  & 467.5 & 7.91$\times$ & 1.23  & 96.23 \scriptsize{$\pm$ 0.06} \\
SIGN         & 352.8 & 5.97$\times$ & 2.17 & 96.12 \scriptsize{$\pm$ 0.05} \\
\hdashline
\textbf{\sg}                  & 59.1 &   & 1.51 & 96.01 \scriptsize{$\pm$ 0.03} \\
\bottomrule
\end{tabular}
\end{small}
\end{center}
\vskip -0.15in
\end{table}

\begin{table}[tb]
\caption{Runtime, memory \& accuracy results on \textit{ogbn-products}. }
\label{tab.products_gpu}
\vskip 0.1in
\begin{center}
\begin{small}
\begin{tabular}{lccc}
\toprule
  & \makecell[c]{Avg. Time \\ \lbrack ms\rbrack} & \makecell[c]{Max Mem \\ \lbrack GB\rbrack} & \makecell[c]{Test  Acc. \\ \lbrack\%\rbrack} \\
\midrule
GraphSAGE-mb & 2.42 & 7.29 & 79.25  \scriptsize{$\pm$ 0.22}\\
\textbf{\sg} & 1.49 & 4.43 & 79.97  \scriptsize{$\pm$ 0.23}\\

Improvement & 1.62$\times$ & 1.65$\times$ & 0.72 $(\uparrow)$ \\
\midrule
ClusterGCN & 2.90 & 6.59 & 78.51  \scriptsize{$\pm$ 0.33} \\
\textbf{\sg} & 1.97 & 3.36 & 79.34  \scriptsize{$\pm$ 0.41} \\

Improvement & 1.47$\times$ & 1.96$\times$ & 0.83 $(\uparrow)$ \\
\midrule
GraphSAINT-E & 0.30 & 7.16 & 79.54  \scriptsize{$\pm$ 0.27} \\
\textbf{\sg} & 0.28 & 3.92  & 80.20 \scriptsize{$\pm$ 0.23} \\

Improvement & 1.07$\times$ & 1.83$\times$ & 0.66 $(\uparrow)$ \\
\bottomrule
\end{tabular}
\end{small}
\end{center}
\vskip -0.15in
\end{table}

\begin{table}[!tb]
\caption{Runtime, memory \& accuracy results on \textit{ogbn-proteins}.}
\label{tab.proteins_gpu}
\vskip 0.1in
\begin{center}
\begin{small}
\begin{tabular}{lcccc}
\toprule
  & \makecell[c]{Avg. \\ Time \\ \lbrack sec\rbrack} & \makecell[c]{Max \\ Mem \\ \lbrack GB\rbrack} & \makecell[c]{Valid\\ Acc. \\ \lbrack\%\rbrack} & \makecell[c]{Test \\ Acc. \\ \lbrack\%\rbrack} \\
\midrule
GAT      & 6.20  & 10.77 & 92.08 \scriptsize{$\pm$ 0.08} & 87.20 \scriptsize{$\pm$ 0.17}\\
\textbf{\sg}      & 4.09  & 6.22  & 92.51 \scriptsize{$\pm$ 0.08} & 86.41 \scriptsize{$\pm$ 0.18} \\
Improvement & 1.52$\times$ & 1.73$\times$ & 0.43 $(\uparrow)$ & 0.79 $(\downarrow)$\\
\bottomrule
\end{tabular}
\end{small}
\end{center}
\vskip -0.15in
\end{table}

We also combine \sg with popular mini-batch training methods. For the largest \textit{ogbn-products} dataset, we implement \sg together with three competitive GNN baselines, namely GraphSAGE, ClusterGCN and GraphSAINT. The results are summarized in Table \ref{tab.products_gpu}. \sg provides a better solution that leads to runtime speedup, memory reduction and even a slightly increased test accuracy for all three methods. We hypothesize that the graph partitioning eliminates some task-irrelevant edges in the original graph, and thus leads to better generalization of GNNs. 

Table \ref{tab.proteins_gpu} provides results on the dense \textit{ogbn-proteins} graph. When it comes to training GNNs on dense graphs, memory poses a significant challenge due to the neighborhood expansion problem. The results show that GAT suffers from considerable memory usage. In contrast, \sg effectively alleviates the issue with $1.52\times$ runtime speedup and 1.73$\times$ memory reduction. Due to space limitations, our results on \textit{Flickr} data are presented in Appendix \ref{apsec.exp}.

\begin{table*}[!tb]
\caption{Average training time per epoch [sec] of \sg compared with GraphSAINT and GCN on \textit{Flickr} and \textit{ogbn-arxiv} data. We record the training time on five platforms with CPU models listed. OOM denotes Out Of Memory. We note that training a GCN on Raspberry Pi 3B is infeasible since it exceeds memory, while \sg still works.}
\label{tab.runtime_cpu1}
\vskip 0.1in
\begin{center}
\begin{small}
\begin{tabular}{l|lcccccc}
\toprule
\multirow{2}{*}{Dataset}& & RPi 3B & Jetson & Phone & Laptop & Desktop-CPU & Desktop-GPU \\
~ & & Cortex-A53 & Cortex-A57 & SDM-845 & i5-8279U & Zen2 3970X & RTX3090 \\
\midrule
\multirow{3}{*}{\textit{Flickr}} & GraphSAINT-N  & 104.1 & 16.86& 7.67 & 2.86 & 1.48 & 0.097 \\
~ & \textbf{\sg} &   48.2 & 7.61 & 3.54 & 1.21 & 0.67 & 0.050 \\
~ & Speedup & $2.16\times$ & $2.22\times$ & $2.17\times$ & $2.36\times$ & $2.24\times$ & $1.94\times$ \\
\midrule
\midrule
\multirow{3}{*}{\textit{ogbn-arxiv}} & GCN & OOM &  28.10 & 21.96  & 13.80 & 5.16 & 0.027 \\
~ & \textbf{\sg} & 501.59 & 18.39 & 13.33 & 6.51 & 2.71 & 0.016 \\
~ & Speedup & - & $1.53\times$ & $1.65\times$ & $2.12\times$ & $1.91\times$ & $1.69\times$ \\
\bottomrule
\end{tabular}
\end{small}
\end{center}
\vskip -0.15in
\end{table*}


\subsubsection{Evaluations on mobile and edge devices}

Following the GPU setting, we proceed to evaluate \sg on mobile and edge devices with CPUs.

\begin{table}[!t]
\caption{Runtime comparison against baseline methods on three large datasets. Average training time per epoch [sec] is reported. Baseline refers to GraphSAGE for \textit{Reddit} and \textit{ogbn-products}. GAT is the baseline for \textit{ogbn-proteins}.}
\label{tab.runtime_cpu2}
\vskip 0.1in
\begin{center}
\begin{small}
\begin{tabular}{lccc}
\toprule
 & \textit{Reddit} & \makecell[c]{ogbn- \\ products} & \makecell[c]{ogbn- \\ proteins}\\
 \midrule
Baseline & 2.02 & 170.75 & 269.70 \\ 
\textbf{\sg} & 0.88 & 77.05 & 142.7 \\
Speedup & $2.30\times$ & $2.22\times$ & $1.89\times$ \\
\bottomrule
\end{tabular}
\end{small}
\end{center}
\vskip -0.15in
\end{table}

\begin{figure}[!b]
  \centering
  \begin{minipage}[b]{0.23\textwidth}
    \includegraphics[width=\textwidth]{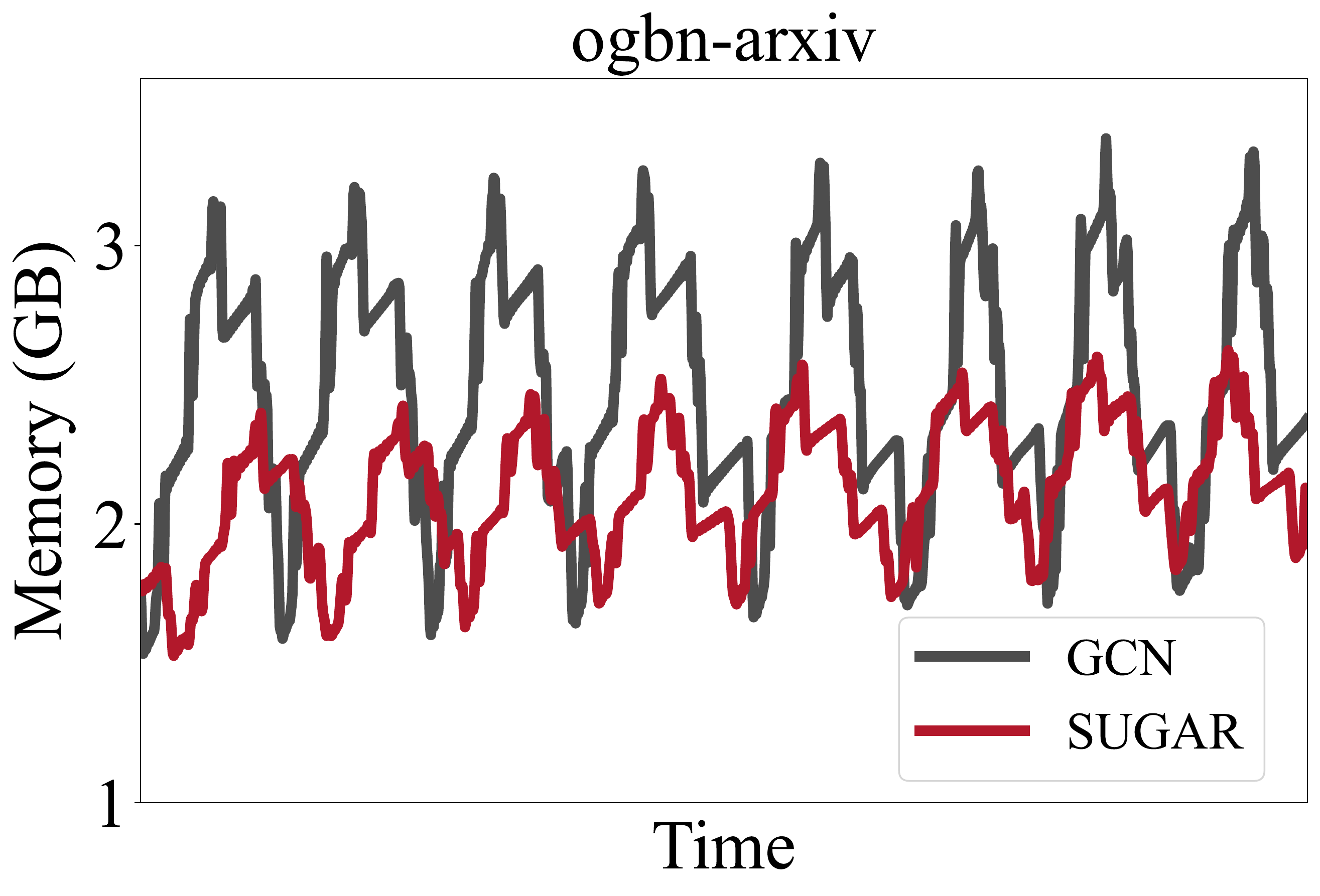}
  \end{minipage}
  \begin{minipage}[b]{0.23\textwidth}
    \includegraphics[width=\textwidth]{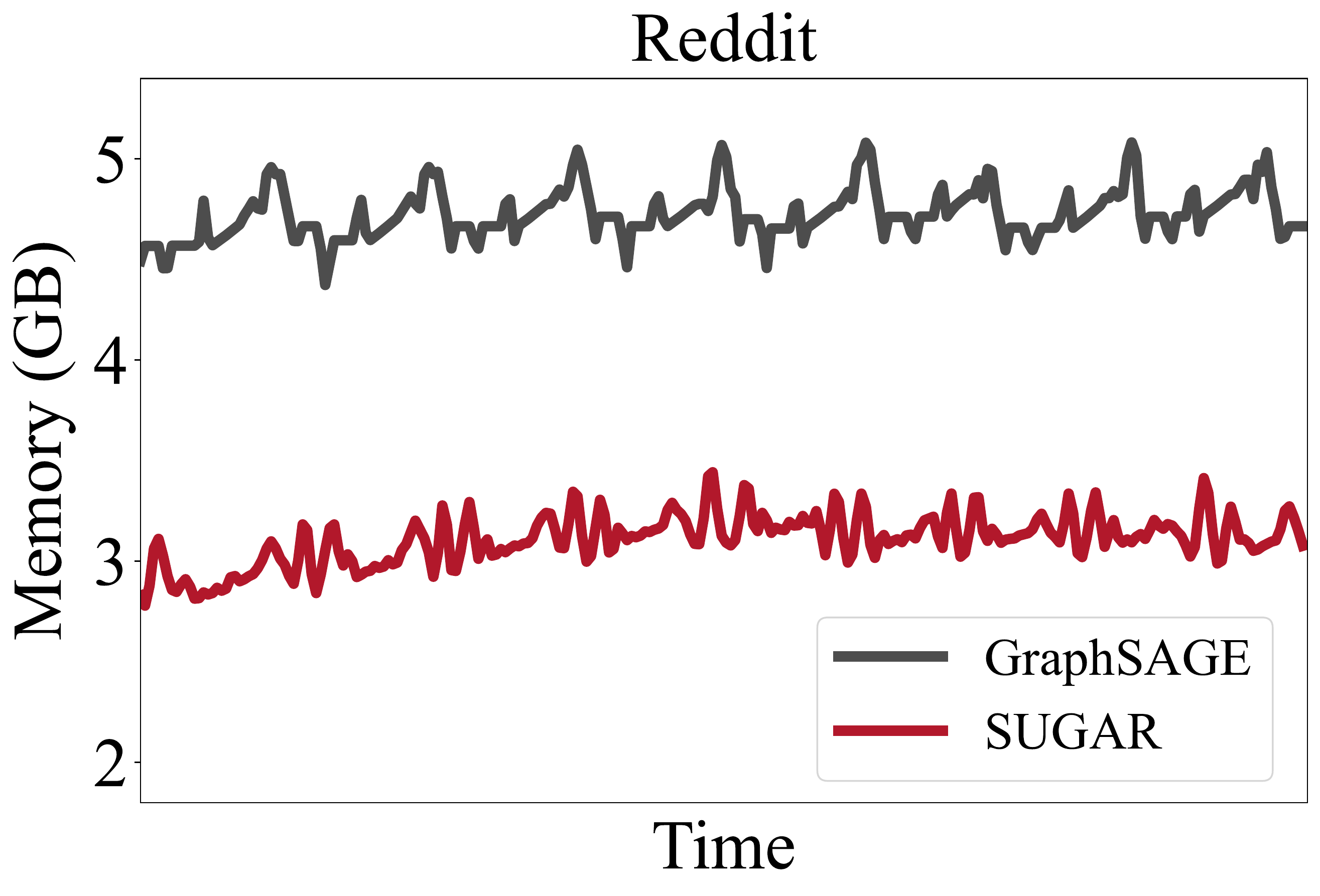}
  \end{minipage}
  \caption{Memory variation during training GNNs on Desktop-CPU for \textit{ogbn-arxiv} and \textit{Reddit}. For \textsc{SUGAR}, we plot the memory variation of the device that consumes most memory. More results are provided in Appendix \ref{apsec.exp}.}
  \label{fig.mem}
\end{figure}

\textbf{Training Time.} Table \ref{tab.runtime_cpu1} presents the average training time per epoch of \sg compared with baselines on the \textit{Flickr} and \textit{ogbn-arxiv} datasets. Due to the relative small size of these two datasets, we are able to train GNNs on all five hardware devices, ranging from a Raspberry Pi 3B, to a desktop equipped with high-performance CPUs. We also list the runtime on GPUs in the last column for easy comparison. 

From Table \ref{tab.runtime_cpu1}, we can see that \sg demonstrates consistent speedup across all platforms, achieving over $2\times$ and $1.5\times$ speedup on the \textit{Flickr} and \textit{ogbn-arxiv} datasets, respectively. In addition, training a GCN on the Raspberry Pi 3B fails due to running out of memory, while \sg demonstrates good memory scalability and hence it can be used with such a device with a limited memory budget (\ie, 1GB in this case). This also holds true for the \textit{Reddit} dataset: \sg provides a feasible solution for local training on the Jetson Nano (time per epoch is 50.27s), while other baselines can not work due to large memory requirements. 

Thus, for the other three datasets, we compare the runtime on Desktop-CPU and report our results in Table \ref{tab.runtime_cpu2}. We also observe consistent speedup across all datasets: \sg nearly halves the training time in all three cases.


\textbf{Memory Usage}. We compare the memory usage of \sg against GNN baselines on a CPU setting. Figure \ref{fig.mem} illustrates the resident set size (RSS) memory usage during training on \textit{ogbn-arxiv} and \textit{Reddit}. Results on the other datasets are provided in Appendix \ref{apsec.exp}. It is evident that our proposed \sg achieves substantial memory reductions. We emphasize that memory plays a critical role in GNN training. In the context of devices with limited resources, the situation is more severe since the graph dataset is already big and loading the full dataset may not be possible. By adopting subgraph-level training, \sg effectively alleviates the problem.

Finally, we present a case study of \sg on NVIDIA Jetson Nano in Appendix \ref{apsec.exp} to demonstrate the applicability of \sg to edge devices.


\subsection{Scalability Analysis}

So far we have demonstrated the great performance of \sg with two available devices. A natural follow-up question is, \textit{how does SUGAR perform on more devices, \ie, device number $K>2$}. Due to space limitations, we provide a detailed scalability analysis in Appendix \ref{apsec.scale}. In short, we observe that there exists a tradeoff between training scalability and performance. With more number of devices available, \sg leads to greater training speedup and smaller memory usage at the cost of slightly degraded performance.  Thus, this provides a feasible solution in extremely resource-limited scenarios while general GNN training methods are not applicable.
Finally, we study the influence of batch sizes on computational efficiency and memory scalability on \sg when compared with mini-batch training algorithms. Our results of a detailed analysis are provided in Appendix \ref{apsec.scale}.

%% file: 5.conclusion.tex
\section{Conclusion}

We have proposed \textsc{SUGAR}, an efficient GNN training method that improves training scalability with multiple devices. \sg can reduce computation, memory and communication costs during training through two key contributions: (1) a novel graph partitioning strategy with memory budgets and graph topology taken into consideration; (2) subgraph-level local GNN training. We have provided a theoretical analysis and conducted extensive experiments to demonstrate the efficiency of \textsc{SUGAR} with real datasets and hardware devices.

\section*{Acknowledgements}
We thank Zhengqi Gao (\textit{EECS Dept., MIT}) and Xinyuan Cao (\textit{CS Dept., Gatech}) for their contributions in the theoretical derivations of this work. In particular, Zhengqi Gao proposed key proofs for Theorem 2 and Proposition 1 in the paper. The authors want to give gratitude to them for their invaluable help and support.



%% file: 6.appendix.tex
\onecolumn
\setcounter{section}{0}
\section{Theoretical Analysis (Section \ref{sec.method}, Page 3-4 in the paper)}\label{apsec.proof1}
In this section, we provide details of the following theoretical results used in the main paper:

(1) \textbf{Lemma \ref{lem.2act}}: For a multi-layer GCN with fixed weights, the error of the activations of the SG estimator are bounded.

(2) \textbf{Lemma \ref{lem.grad}}: For a multi-layer GCN with fixed weights, the error of the gradients of the SG estimator are bounded.

(3) \textbf{Theorem \ref{thm.converge}}: With high probability gradient descent training with the approximated gradients by the SG estimator can converge to a local minimum.

The proof builds on \cite{chen2017stochastic}, but with different assumptions. More precisely, while \cite{chen2017stochastic} assume that model weights change slowly during training, our theoretical analysis is based on the difference in the adjacency matrices produced by graph partitioning. 

\subsection{Notations}
Let $[L]=\{1,...,L\}$. The infinity norm of a matrix is defined as $\norm{A} = \max_{i,j} |A_{i,j}|$. By Proposition B in \cite{chen2017stochastic}, we know that:
\begin{enumerate}
    \item $\norm{AB} \leq col(A)\norm{A}\norm{B}$ 
    \item $\norm{A\circ B} \leq \norm{A}\norm{B}$
    \item $\norm{A+B} \leq \norm{A} + \norm{B}$
\end{enumerate}
where $col(A)$ represents the number of columns of matrix $A$ and $\circ$ is the element-wise product. We define $\eta$ to be the maximum number of columns we can possibly encounter in the proof.

We review some notations defined in the main text.\footnote{Some equations in the Appendix may have different numbers from the main paper.} Our proposed estimator is denoted by SG.
The propagation rule of a $l$-th layer GCN with the exact estimator is given by:
\begin{equation}
	Z^{(l+1)} = A^{norm}H^{(l)}W^{(l)}, H^{(l+1)} = \sigma(Z^{(l+1)})
	\label{eq.exact}
\end{equation}
Similarly, the propagation rule of a $l$-th layer GCN with the SG estimator is given by:
\begin{equation}
	Z_{SG}^{(l+1)} = A_{SG}^{norm}H_{SG}^{(l)}W^{(l)}, H_{SG}^{(l+1)} =  \sigma(Z_{SG}^{(l+1)})
	\label{eq.sgestimate}
\end{equation}
where $\sigma$ represents an activation function, $A^{norm}$ denotes the normalized version of $A$, \ie, $A^{norm}=\hat D^{-1/2}\hat A \hat D^{1/2}, \hat A = A + I_N, \hat D_{ii}=\sum_j\hat A_{ij}$ and $I_N$ is an $N$-dimensional identity matrix.$H^{(l)}$ and $H_{SG}^{(l)}$ denote node representations in the $l$-th layer produced by the exact GCN and SG estimator, respectively. $W^{(l)}$ represents the weight matrix in layer $l$. Note that while we write Equation \ref{eq.sgestimate} in a compact matrix form, in real implementation, the training process is distributed across $K$ devices.

Recall that $A_{SG}$ is a block-diagonal matrix produced by the graph partitioning module that serves as an approximation of $A$. Before training, we run graph partitioning for $M$ times to obtain a sample average, \ie, $A_{SG}^{norm} = \frac{1}{M} \sum_{m=1}^M{A_{SG, m}^{norm}}$. Let $\epsilon=\norm{A_{SG}^{norm}-A^{norm}}$ denote the error in approximating $A^{norm}$ with $A_{SG}^{norm}$. For simplicity, we will omit the superscript $norm$ from now on. 

The model parameters at training epoch $t$ are denoted by $W_t$. For $W$ at a given time point (\ie, fixed model weights), we omit the subscript in the proof. Let $W_*$ denote the optimal model weights. $\nabla \cl(W) = \frac{1}{N}\sum_{i=1}^N \frac{\partial f(y_i, z_i^{(L)})}{\partial W}$ and $\nabla \cl_{SG}(W) =\frac{1}{N}\sum_{i=1}^N \frac{\partial f(y_i, z_{SG, i}^{(L)})}{\partial W}$ represent the gradients of the exact GCN and SG estimator with respect to model weights $W$, respectively. $f(\cdot, \cdot)$ is the objective function (\eg, cross entropy for node classification tasks).



\subsection{Activations of Multi-layer GCN}
\subsubsection{Single-layer GCN}
Proposition \ref{prop.1gcn} states that for a single-layer GCN, (1) the outputs are bounded if the inputs are bounded, (2) if the difference between the input of the SG estimator and the exact GCN is small, then the output of the SG estimator is close to the output of the exact GCN.
\begin{prop}
For a one-layer GCN, if the activation function $\sigma(\cdot)$ is $\rho$-Lipschitz and $\sigma(0) = 0$, for any input matrices $A$, $A_{SG}$, $X$, $X_{SG}$ and any weight matrix $W$ that satisfy:
\begin{enumerate}
    \item All the matrices are bounded by $\beta$: $\norm{A}\leq \beta$, $\norm{A_{SG}}\leq \beta$, $\norm{X} \leq \beta$, $\norm{X_{SG}} \leq \beta$ and $\norm{W} \leq \beta$,
    \item The differences between inputs are bounded: $\norm{X_{SG}-X} \leq \alpha \epsilon$, where $\epsilon = \norm{A_{SG}-A}$.
    
\end{enumerate}
Then, there exist $B$ and $C$ that depend on $\rho$, $\eta$ and $\beta$, \st,

\begin{enumerate}
    \item The outputs are bounded: $\norm{H} \leq B$ and $\norm{H_{SG}} \leq B$, 
    \item The differences between outputs of the SG estimator and the exact estimator are bounded: $\norm{Z_{SG}-Z} \leq C(1+\alpha)\epsilon$ and $\norm{H_{SG}-H} \leq C(1+\alpha)\epsilon$.
\end{enumerate}
\label{prop.1gcn}
\end{prop}

\textit{Proof.}
We know that $\norm{Z} = \norm{A X W} \leq \eta^2 \norm{A}\norm{X}\norm{W} \leq \eta^2 \beta^3 $. By Lipschitz continuity of $\sigma(\cdot)$, $\norm{\sigma(Z)-\sigma(0)} \leq \rho \eta^2 \beta^3 $ and we have $\norm{\sigma(Z)} \leq \rho\eta^2\beta^3$. Thus $\norm{H} \leq D$, where $B=\max\{ \eta^2\beta^3, \rho\eta^2\beta^3 \}$. Similarly, $\norm{H_{SG}} \leq B$.

We proceed to show that the differences between outputs are bounded below:
\begin{equation}
\begin{aligned}
    \norm{Z_{SG}-Z} & = \norm{A_{SG}X_{SG}W - AXW} \\
    & \leq \eta \norm{W} \norm{A_{SG}X_{SG}-AX} \\
    & \leq \eta \beta (\norm{A_{SG}(X_{SG}-X)} + \norm{X(A_{SG}-A)}) \\
    & \leq \eta \beta (\eta \beta \alpha \epsilon + \eta \beta \epsilon) \\
    & = (1+\alpha) \eta^2 \beta^2  \epsilon
    \label{eq.zdif}
\end{aligned}
\end{equation}

By Lipschitz continuity of $\sigma(\cdot)$, we have $\norm{H_{SG}-H_{t}} \leq \rho (1+\alpha)\eta^2 \beta^2 \epsilon$. Choose $C=\max \{ (1+\alpha)\eta^2 \beta^2, \rho(1+\alpha) \eta^2 \beta^2\}$, and the proof is complete.

\subsubsection{Multi-layer GCN}
The following lemma relates the approximation error in activations (\ie, $\norm{H_{SG}^{(l)}-H^{(l)}}$) with the approximation error in input adjacency matrices (\ie, $\epsilon=\norm{A_{SG}-A}$). 
\setcounter{lem}{0} 
\begin{lem}
For a multi-layer GCN with fixed model weights, given a (fixed) graph dataset, assume that:
\begin{enumerate}
    \item $\sigma(\cdot)$ is $\rho$-Lipschitz and $\sigma(0) = 0$,
    \item The inputs are bounded by $\beta$: $\norm{A}\leq \beta$, $\norm{A_{SG}}\leq \beta$, $\norm{X} \leq \beta$,
    \item The model weights in each layer are bounded by $\beta$: $\norm{W^{(l)}} \leq \beta, \forall l \in [L]$.
\end{enumerate}

Then, there exist $B$ and $C$ that depend on $\rho$, $\eta$ and $\beta$, \st,

\begin{enumerate}
    \item $\norm{H^{(l)}} \leq B, \norm{H_{SG}^{(l)}} \leq B, \ \forall l \in [L-1]$,
    \item $\norm{Z_{SG}^{(l)}-Z^{(l)}} \leq C\epsilon,\  \forall l \in [L]$ and $\norm{H_{SG}^{(l)}-H^{(l)}} \leq C\epsilon,\ \forall l \in [L-1]$.
\end{enumerate}

\label{lem.2act}
\end{lem}

\textit{Proof.} 
Applying Proposition \ref{prop.1gcn} to each layer of the GCN proves that $H^{(l)}$ and $H_{SG}^{(l)}$ are bounded for each layer $l$.

For the first layer of GCN, by Proposition \ref{prop.1gcn} and input conditions, we know that there exists $C^{(1)}$ that satisfies:
\begin{equation*}
    \norm{Z_{SG}^{(1)}-Z^{(1)}} \leq C^{(1)}\epsilon, \quad \norm{H_{SG}^{(1)}-H^{(1)}} \leq C^{(1)}\epsilon
\end{equation*}
Note that for the first layer, the node feature matrix of the SG estimator and exact GCN are identical, \ie, $X_{SG} = X$; this yields $\alpha=0$ in Equation \ref{eq.zdif}. 
Let $\hat C^{(1)}=C^{(1)}$. Next, we apply Proposition \ref{prop.1gcn} to the second layer of GCN; there exists $C^{(2)}$ that satisfies: \begin{equation*}
    \norm{Z_{SG}^{(2)}-Z^{(2)}} \leq C^{(2)}(1+\hat C^{(1)})\epsilon, \quad \norm{H_{SG}^{(2)}-H^{(2)}} \leq C^{(2)}(1+\hat C^{(1)})\epsilon
\end{equation*}

Let $\hat C^{(2)}=C^{(2)}(1+\hat C^{(1)})$. By applying Proposition \ref{prop.1gcn} to the subsequent layer of GCN repetitively, we have $\hat C^{(l+1)}=C^{(l+1)}(1+\hat C^{(l)}), \forall l \in [L-1]$. We choose $C = \max_l \hat C^{(l)}$ and complete the proof.







\subsection{Gradients of Multi-layer GCN}
Lemma \ref{lem.grad} below provides a bound for the difference between gradients of the loss by the SG estimator and the exact GCN (\ie, $\norm{\nabla \cl_{SG}(W)- \nabla \cl(W)}$). Intuitively, the gradient difference is small if the approximation error in input adjacency matrices (\ie, $\epsilon$) is small. 
\begin{lem}
For a multi-layer GCN with fixed model weights, given a (fixed) graph dataset, assume that:
\begin{enumerate}
    \item $\frac{\partial f(y, z)}{\partial z}$ is $\rho$-Lipschitz and $\norm {\frac{\partial f(y, z)}{\partial z}} \leq \beta$,
    \item $\sigma (\cdot)$ is $\rho$-Lipschitz, $\sigma(0) = 0$ and $\norm{\sigma'(\cdot)} \leq \beta$,
    \item $\norm{A}\leq \beta$, $\norm{A_{SG}}\leq \beta$, $\norm{X} \leq \beta$, $\norm{W^{(l)}} \leq \beta, \forall l \in [L]$.
\end{enumerate}

Then, there exists $C$ that depends on $\rho$, $\eta$ and $\beta$, \st, $\norm{\nabla \cl_{SG}(W)- \nabla \cl(W)} \leq C \epsilon$.
\label{lem.grad}
\end{lem}
\textit{Proof.} We begin by proving the following statements:

{\it
If the above assumptions hold, then there exist $C$ and $D$ that depends on $\rho$, $\eta$ and $\beta$, s.t.,
\begin{enumerate}
    \item The gradients with respect to the activations of each layer of the SG estimator are close to be unbiased: 
    \begin{equation}
   \norm{\frac{\partial f}{\partial Z_{SG}^{(l)}} - \frac{\partial{f}}{{\partial Z^{(l)}}}} \leq C \epsilon, \quad \forall l \in [L]
   \label{eq.gradz}
  \end{equation}
  
   \item The gradients above are bounded: 
   \begin{equation}
       \norm{\frac{\partial f}{\partial Z_{SG}^{(l)}}} \leq D \beta, \quad \norm{\frac{\partial f}{\partial Z^{(l)}}} \leq D \beta, \quad \forall l \in [L]
       \label{eq.gradnorm}
   \end{equation} 

\end{enumerate}}

We prove these statements by induction. First we show that Equations \ref{eq.gradz} and \ref{eq.gradnorm} hold true for the final layer of GCN (\ie, $l = L$). By Assumption 1 and Lemma \ref{lem.2act}, we know that there exists $\hat C$ that satisfies:

\begin{equation}
   \norm{\frac{\partial f}{\partial Z_{SG}^{(L)}} - \frac{\partial{f}}{{\partial Z^{(L)}}}} \leq \rho \norm{Z_{SG}^{(L)}-Z^{(L)}} \leq \rho \hat C \epsilon
\end{equation}

Let $C^{(L)} = \rho \hat C$ and $D^{(L)} = 1$. Next, suppose the statements hold for layer $l+1$, \ie, there exist $C^{(l+1)}$ and $D^{(l+1)}$ that satisfy:
\begin{equation}
    \begin{aligned}
        \norm{\frac{\partial f}{\partial Z_{SG}^{(l+1)}} - \frac{\partial{f}}{{\partial Z^{(l+1)}}}} & \leq C^{(l+1)} \epsilon, \quad
   \norm{\frac{\partial f}{\partial Z_{SG}^{(l+1)}}} \leq D^{(l+1)} \beta, \quad   \norm{\frac{\partial f}{\partial Z^{(l+1)}}} \leq D^{(l+1)} \beta \label{eq.induction_l1}
    \end{aligned}
\end{equation}
We derive the gradients of the objective function with respect to activations in layer $l$ by chain rule:  
\begin{equation}
\begin{aligned}
    \norm{\frac{\partial f}{\partial Z^{(l)}}} & = \norm{\sigma'(Z^{(l)}) \circ \frac{\partial f}{\partial H^{(l)}}} \\
    & = \norm{\sigma'(Z^{(l)}) \circ A^T \ \frac{\partial f}{\partial Z^{(l+1)}} \  {W^{(l)}}^T} \\
    & \leq \eta^2 \norm{\sigma'(Z^{(l)})} \norm{A} \norm{\frac{\partial f}{\partial Z^{(l+1)}}} \norm{W^{(l)}} \\
    & \leq \eta^2 \beta^4 D^{(l+1)} 
\end{aligned}
\end{equation}
Thus, we know that $ \norm{\frac{\partial f}{\partial Z^{(l)}}} \leq D^{(l)} \beta$. Similarly, $ \norm{\frac{\partial f}{\partial Z_{SG}^{(l)}}} \leq D^{(l)} \beta$, where $D^{(l)}=\eta^2 \beta^3 D^{(l+1)}$. 

We proceed to derive the error of the gradients by the SG estimator in layer $l$:
\begin{equation}
    \begin{aligned}
        \norm{\frac{\partial f}{\partial Z_{SG}^{(l)}} -  \frac{\partial{f}}{{\partial Z^{(l)}}}} & = \norm{\sigma'(Z_{SG}^{(l)}) \circ A_{SG}^T \ \frac{\partial f}{\partial Z_{SG}^{(l+1)}} \  {W^{(l)}}^T - \sigma'(Z^{(l)}) \circ A^T \ \frac{\partial f}{\partial Z^{(l+1)}} \  {W^{(l)}}^T} \\
        & \leq \eta \norm{W^{(l)}} \norm{\sigma'(Z_{SG}^{(l)}) \circ A_{SG}^T \ \frac{\partial f}{\partial Z_{SG}^{(l+1)}} - \sigma'(Z^{(l)}) \circ A^T \ \frac{\partial f}{\partial Z^{(l+1)}}} \\
        & \leq \underbrace{\eta \beta \norm{ (\sigma'(Z_{SG}^{(l)})-\sigma'(Z^{(l)})) \circ A_{SG}^T \frac{\partial f}{\partial Z_{SG}^{(l+1)}}}}_{(*)} + \underbrace{\eta \beta \norm{\sigma'(Z^{(l)}) \circ A_{SG}^T (\frac{\partial f}{\partial Z_{SG}^{(l+1)}} - \frac{\partial f}{\partial Z^{(l+1)}})}}_{(**)} \\
        & + \underbrace{ \eta \beta \norm{\sigma'(Z^{(l)}) \circ (A_{SG}^T-A^T) \frac{\partial f}{\partial Z^{(l+1)}}}}_{(***)}
    \end{aligned}
    \label{eq.induction_l}
\end{equation}

By Assumption 2 and Lemma \ref{lem.2act}, we know that there exists $\hat C$ such that $\norm{ \sigma'(Z_{SG}^{(l)})-\sigma'(Z^{(l)})} \leq \rho \hat C \epsilon$. From Equation \ref{eq.induction_l1}, we have:
\begin{equation}
\begin{aligned}
  \text{(*) in Equation (\ref{eq.induction_l})} & \leq \eta^2 \beta \norm{ \sigma'(Z_{SG}^{(l)})-\sigma'(Z^{(l)})} \norm{A_{SG}} \norm{\frac{\partial f}{\partial Z_{SG}^{(l+1)}}}  \\
    & \leq \eta^2 \beta \cdot \rho \hat C \epsilon \cdot \beta \cdot D^{(l+1)} \beta \\
    & = (\eta^2 \beta^3  \rho \hat C D^{(l+1)}) \epsilon \\
    \text{(**) in Equation (\ref{eq.induction_l})} & \leq \eta^2 \beta \norm{\sigma'(Z^{(l)})} \norm{A_{SG}} \norm{\frac{\partial f}{\partial Z_{SG}^{(l+1)}} - \frac{\partial f}{\partial Z^{(l+1)}}} \\
    & \leq \eta^2 \beta \cdot \beta \cdot \beta \cdot C^{(l+1)} \epsilon \\
    & = (\eta^2 \beta^3 C^{(l+1)}) \epsilon \\
    \text{(***) in Equation (\ref{eq.induction_l})} & \leq \eta^2 \beta \norm{\sigma'(Z^{(l)})} \norm{A_{SG}^T-A^T} \norm{\frac{\partial f}{\partial Z^{(l+1)}}} \\
    & \leq \eta^2 \beta \cdot \beta \cdot \epsilon \cdot D^{(l+1)} \beta \\
    & = (\eta^2 \beta^3 D^{(l+1)}) \epsilon
\end{aligned}
\end{equation}

Therefore, $
     \norm{\frac{\partial f}{\partial Z_{SG}^{(l)}} -  \frac{\partial{f}}{{\partial Z^{(l)}}}} \leq C^{(l)} \epsilon
$, where $C^{(l)} = \eta^2 \beta^3 [(\rho \hat C+1) D^{(l+1)} + C^{(l+1)}]$. By induction, Equations \ref{eq.gradz} and \ref{eq.gradnorm} hold true.

Next, we show below that there exists $C$ that depends on $\rho$, $\eta$ and $\beta$, \st,
\begin{equation}
    \begin{aligned}
    \norm{\frac{\partial f}{\partial W_{SG}^{(l)}} -  \frac{\partial{f}}{{\partial W^{(l)}}}} \leq C \epsilon,\quad \forall l \in [L]
    \end{aligned}
    \label{gradw}
\end{equation}

By backpropagation rule we derive that $\frac{\partial f}{\partial W^{(l)}} = (AH^{(l)})^T \frac{\partial{f}}{{\partial Z^{(l)}}}$. By Lemma \ref{lem.2act}, we know that $H_{SG}^{(l)}$ is bounded by some $\hat B$ and $\norm{H_{SG}^{(l)}-H^{(l)}} \leq \tilde C \epsilon$ hold for some $\tilde C$. From the previous proof, we know that there exists $\hat C$ and $\hat D$, \st, Equations \ref{eq.gradz} and \ref{eq.gradnorm} hold; thus, we have:
\begin{equation} 
    \begin{aligned}
     \norm{ \frac{\partial f}{\partial W_{SG}^{(l)}} - \frac{\partial f}{\partial W^{(l)}} }
     & \leq \norm{(A_{SG}H_{SG}^{(l)})^T \frac{\partial{f}}{{\partial Z_{SG}^{(l+1)}}} - (AH^{(l)})^T \frac{\partial{f}}{{\partial Z^{(l+1)}}}} \\
     & \leq \norm{(A_{SG}H_{SG}^{(l)})^T (\frac{\partial f}{\partial {Z_{SG}^{(l+1)}}}-\frac{\partial f}{\partial {Z^{(l+1)}}}) } + \norm{((A_{SG}H_{SG}^{(l)})^T-(AH^{(l)})^T) \frac{\partial f}{\partial {Z^{(l+1)}}}} \\
     & \leq \eta^2 \beta \cdot \hat B \cdot \hat C \epsilon + \eta \norm{A_{SG}H_{SG}^{(l)}-AH^{(l)}} \hat D \beta \\
     & \leq \eta^2 \beta \hat B \hat C \epsilon + \eta \beta \hat D \norm{A_{SG}(H_{SG}^{(l)}-H^{(l)})} + \eta \beta \hat D \norm{(A_{SG}-A)H^{(l)}} \\
     & \leq \eta^2 \beta \hat B \hat C \epsilon + \eta \beta \hat D \cdot \eta \beta \cdot \tilde C \epsilon + \eta \beta \hat D \cdot \eta \epsilon \cdot \hat B \\
     & = \eta^2 \beta (\hat B \hat C + \beta \tilde C \hat D + \hat B \hat D) \epsilon
    \end{aligned}
\end{equation}
Therefore, Equation \ref{gradw} holds, where $C=\eta^2 \beta (\hat B \hat C + \beta \tilde C \hat D + \hat B \hat D)$.

Finally, we have:  $\norm{\nabla \cl_{SG}(W)- \nabla \cl(W)} \leq C \epsilon$, and the proof is complete.

\subsection{Convergence Analysis}
\setcounter{thm}{0}

\begin{thm}
Assume that: 
\begin{enumerate}
    \item The loss function $\mathcal L(W)$ is $\rho$-smooth, \ie, $|\mathcal L(W_2) - \mathcal L(W_1) - \langle \mathcal L(W_1), W_2-W_1 \rangle | \leq \frac{\rho}{2} \fnorm{W_2-W_1}^2, \forall W_1, W_2$, where $\langle A,B \rangle = tr(A^TB)$ denotes the inner product of matrix $A$ and $B$,
    \item The gradients of the loss $\nabla \cl(W)$ and $\nabla \cl_{SG}(W)$ are bounded by $G$ for any choice of $W$,
    \item The gradient of the objective function $\frac{\partial f(y, z)}{\partial z}$ is $\rho$-Lipschitz and bounded,
    \item The activation function $\sigma (\cdot)$ is $\rho$-Lipschitz, $\sigma(0) = 0$ and $\sigma'(\cdot)$ is bounded.
\end{enumerate}

Then there exists $C>0$, \st, $\forall M, T$, for a sufficiently small $\delta$, if we run graph partitioning for $M$ times and run gradient descent for $R \leq T$ epochs (where $R$ is chosen uniformly from $[T]$, the model update rule is $W_{t+1} = W_t - \gamma \nabla \mathcal L_{SG}(W_t)$, step size $\gamma = \frac{1}{\rho\sqrt T}$), we have:
\begin{equation*}
    P(\mathbb E_{R} \fnorm{\nabla \cl (W_R)}^2 \leq \delta) \geq 1-2 \exp \{-2M (\frac{\delta}{2C} - \frac{2\rho [\cl(W_1) - \cl(W_*)] + C - \delta}{2C(\sqrt T -1)})^2 \}
\end{equation*}
\end{thm}

\textit{Proof.} Let $\delta_t = \nabla \cl_{SG}(W_t) - \nabla \cl(W_t)$ denote the differences between gradients at epoch $t$. By $\rho$-smoothness of $\cl(W)$ we know that:
\begin{equation}
\begin{aligned}
    \cl(W_{t+1}) &\leq \cl(W_t) + \langle \nabla \cl(W_t), W_{t+1} - W_t \rangle + \frac{\rho}{2} \gamma^2 \fnorm{\nabla \cl_{SG}(W_t)}^2 \\
   & = \cl(W_t) - \gamma \langle \nabla \cl(W_t), \nabla \cl_{SG}(W_t) \rangle + \frac{\rho}{2} \gamma^2 \fnorm{\nabla \cl_{SG}(W_t)}^2 \\
   & = \cl(W_t) - \gamma \langle \nabla \cl(W_t), \delta_t \rangle - \gamma \fnorm{\nabla \cl(W_t)}^2 + \frac{\rho}{2} \gamma^2 [\fnorm{\delta_t}^2 + \fnorm{\nabla \cl(W_t)}^2 + 2 \langle \delta_t, \nabla \cl(W_t)\rangle] \\
   & = \cl(W_t) - (\gamma-\rho\gamma^2) \langle \nabla \cl(W_t), \delta_t \rangle - (\gamma-\frac{\rho}{2}\gamma^2) \fnorm{\nabla \cl(W_t)}^2 + \frac{\rho}{2}\gamma^2 \fnorm{\delta_t}^2
\end{aligned}
\label{eq.smooth}
\end{equation}

By Lemma \ref{lem.grad}, we know that at a given time point $t$, there exists $\hat C$ \st, $\delta_t$ is bounded by $\hat C \epsilon$. Therefore,
\begin{equation}
    \begin{aligned}
    & |\langle \nabla \cl(W_t), \delta_t \rangle | \leq \eta \norm{\nabla \cl(W_t)} \norm{\delta_t} \leq \eta G \hat C \epsilon \\
    & \fnorm{\delta_t}^2 \leq \norm{\nabla \cl_{SG}(W_t)}^2 + \norm{\nabla \cl(W_t)}^2 \leq 2 G^2 \\
    \end{aligned}
\end{equation}
Let $C = \max\{ \eta G\hat C, 2G^2\}$. Equation \ref{eq.smooth} can be further derived as:
\begin{equation}
    \cl(W_{t+1}) \leq \cl(W_{t}) + (\gamma-\rho\gamma^2) C \epsilon - (\gamma-\frac{\rho}{2}\gamma^2) \fnorm{\nabla \cl(W_t)}^2 + \frac{\rho}{2}C\gamma^2
\end{equation}

By summing up the above inequalities from $t=1$ to $T$ and rearranging the terms, we have:
\begin{equation}
    (\gamma-\frac{\rho}{2}\gamma^2) \sum_t \fnorm{\nabla \cl(W_t)}^2 \leq \cl(W_1) - \cl(W_*) + C T(\gamma-\rho\gamma^2)\epsilon + \frac{\rho}{2}CT\gamma^2
    \label{eq.lwt}
\end{equation}
Dividing both sides of Equation \ref{eq.lwt} by $T(\gamma-\frac{\rho}{2}\gamma^2)$ and choosing $\gamma = \frac{1}{\rho \sqrt T}$ gives us:
\begin{equation}
\begin{aligned}
 \mathbb E_{R} \fnorm{\nabla \cl (W_R)}^2 & \leq 2 \frac{\cl(W_1) - \cl(W_*) + C T(\gamma-\rho\gamma^2)\epsilon + \frac{\rho}{2}CT\gamma^2}{T\gamma(2-\rho\gamma)} \\
    & \leq \frac{2[\cl(W_1) - \cl(W_*)] }{T \gamma} + 2C(1-\rho\gamma) \epsilon + \rho C \gamma \\
    & \leq \frac{2\rho [\cl(W_1) - \cl(W_*)]}{\sqrt T} + 2C(1-\frac{1}{\sqrt T}) \epsilon  + \frac{C} {\sqrt T} \\
    & \leq \frac{2\rho [\cl(W_1) - \cl(W_*)] + C}{\sqrt T} + 2C(1-\frac{1}{\sqrt T}) \epsilon
    \label{eq.converge}
\end{aligned}
\end{equation}

Recall that $\epsilon$ denotes the infinity norm of the error in approximating $A$ through $M$ runs, \ie, $\epsilon = \norm{A_{SG}-A}$. Applying Hoeffding's inequality \cite{glynn2002hoeffding} to the largest element of the matrix $|A_{SG}-A|$ (which are bounded by the intervals $[0, 1]$), we have:
\begin{equation}
    P(\epsilon \geq \delta) \leq 2 \exp(-2M \delta^2),\quad \forall \delta \geq 0
    \label{eq.hoeff}
\end{equation}

Combining the two inequalities above, we have:
\begin{equation}
\begin{aligned}
 P(\mathbb E_{R} \fnorm{\nabla \cl (W_R)}^2 \geq \delta) &\leq P(\frac{2\rho [\cl(W_1) - \cl(W_*)] + C}{\sqrt T} + 2C(1-\frac{1}{\sqrt T}) \epsilon \geq \delta) \\
 & \leq 2 \exp \{-2M (\frac{\delta}{2C} - \frac{2\rho [\cl(W_1) - \cl(W_*)] + C - \delta}{2C(\sqrt T -1)})^2\}
\end{aligned}
\end{equation}
Therefore, for a sufficiently small $\delta$, we have the following inequality for $P(\mathbb E_{R} \fnorm{\nabla \cl (W_R)}^2 \leq \delta)$: 
\begin{equation}
    P(\mathbb E_{R} \fnorm{\nabla \cl (W_R)}^2 \leq \delta) \geq 1-2 \exp \{-2M (\frac{\delta}{2C} - \frac{2\rho [\cl(W_1) - \cl(W_*)] + C - \delta}{2C(\sqrt T -1)})^2 \}
\end{equation}
Theorem \ref{thm.converge} is proved.

\section{Proof of Proposition \ref{prop.ub} (Section \ref{sec.method}, Page 4 in the paper)} \label{apsec.proof2}
By convexity of $f$, using Jensen's inequality \cite{jensen} gives us:
\begin{equation}
   f(y_i, \frac{1}{|\mathcal P_i|} \sum_{k\in \mathcal P_i}g(x_i; W^{\langle k \rangle}) \leq \frac{1}{|\mathcal P_i|} \sum_{k\in \mathcal P_i} f(y_i, g(x_i; W^{\langle k \rangle})) 
\end{equation}

By changing the operation order and regrouping the indices, we further derive:
\begin{equation}
\begin{aligned}
    \frac{1}{N} \sum_{i=1}^N \frac{1}{|\mathcal P_i|} \sum_{k\in \mathcal P_i} f(y_i, z_i) =
    \frac{1}{K} \sum_{k=1}^K \sum_{i\in \mathcal V_k} \frac{1}{|\mathcal P_i|}f(y_i, z_i)
\end{aligned}
\end{equation}

Therefore, 
\begin{equation}
\begin{aligned}
    \cl &\leq \frac{1}{K} \sum_{k=1}^K  \sum_{i\in \mathcal V_k} \frac{1}{|\mathcal P_i|} f(y_i, z_i) \\
& z_i = g(x_i, W^{\langle k \rangle}) \\
\end{aligned}
\end{equation}
Proposition \ref{prop.ub} is proved.

\section{Experimental Setup (Section \ref{sec.exp}, Page 5 in the paper)}\label{apsec.setup}
We evaluate \sg on five node classification datasets, selected from very diverse applications: (1) categorizing types of images based on the descriptions and common properties of online images (\textit{Flickr}); (2) predicting communities of online posts based on user comments (\textit{Reddit}); (3) predicting the subject areas of arxiv papers based on its title and abstract (\textit{ogbn-arxiv}); (4) predicting the presence of protein functions based on biological associations between proteins (\textit{ogbn-proteins}); (5) predicting the category of a product in an Amazon product co-purchasing network (\textit{ogbn-products}). Note that the task of \textit{ogbn-proteins} is multi-label classification, while other tasks are multi-class classification.


We include the following GNN architectures and SOTA GNN training algorithms for comparison: 
\begin{itemize}
\item GCN \cite{gcn}: Full-batch Graph Convolutional Networks.
\item GraphSAGE \cite{graphsage}: An inductive representation learning framework that efficiently generates node embeddings for previously unseen data. 
\item GAT \cite{gat}: Graph Attention Networks, a GNN architecture that leverages masked self-attention layers.
\item SIGN \cite{sign}: Scalable Inception Graph Neural Networks, an architecture using graph convolution filters of different size for efficient computation.
\item ClusterGCN \cite{clustergcn}: A mini-batch training technique that partitions the graphs into a fixed number of subgraphs and draws mini-batches from them. 
\item GraphSAINT \cite{graphsaint}: A mini-batch training technique that constructs mini-batches by graph sampling. 
\end{itemize}

Note that our reported experiments involve no training communication among local models, \ie, $K$ local models are trained separately. While training with communication (\eg, maintain a central model to collect gradient updates from local models and do the gradient descent) aligns rigorously with our theoretical analysis in Section \ref{subsec.theory} and is expected to achieve higher accuracy than training without communication, the latter yields great practical benefits as (1) communication is not guaranteed in many real IoT applications, (2) savings in communication costs lead to training speedup, as well as energy reduction per device, which are crucial for model deployment in practice. Moreover, we observed that training without communication already yields satisfactory results compared with its counterpart. For instance, we conducted experiments on \textit{ogbn-arxiv} graph with both settings, and the difference in test accuracy is small (\ie, 0.08\% for a two-device system and 0.47\% for a eight-device system). Thus, we adopt the no training-time communication setting due to its practical value and empirically good performance.
 
\section{More Experimental Results (Section \ref{sec.exp}, Page 6-7 in the paper)}\label{apsec.exp}
Table \ref{tab.flickr_gpu} presents results of \sg integrated with GraphSAINT for three sampler modes (\ie, node, edge, and random walk based samplers) on \textit{Flickr}. Note that the accuracy we obtain (about 50\%) is consistent with results in \cite{graphsaint}. \sg achieves more than $2\times$ runtime speedup and requires less memory than GraphSAINT. Of note, the test accuracy loss is within 1\% in all cases.

\begin{table}[htb]
    \caption{Runtime, memory \& accuracy results on \textit{Flickr}.}
\label{tab.flickr_gpu}
\vskip 0.1in
\begin{center}
\begin{small}
\begin{tabular}{lccc}
\toprule
  & \makecell[c]{Avg. Time  \lbrack ms\rbrack } & \makecell[c]{Max Mem  \lbrack GB\rbrack} & \makecell[c]{Test Acc. \lbrack\%\rbrack} \\
\midrule
GraphSAINT-N & 97.0  & 0.41 & 50.64 \scriptsize{$\pm$ 0.28} \\
\textbf{\sg}      & 49.9  & 0.31 & 50.11 \scriptsize{$\pm$ 0.12} \\
 Improvement& 1.94$\times$ & 1.32$\times$ & 0.53 $(\downarrow)$\\
\midrule
GraphSAINT-E & 71.1  & 0.53 & 50.91 \scriptsize{$\pm$ 0.12} \\
\textbf{\sg}          & 32.6  & 0.41 & 49.96 \scriptsize{$\pm$ 0.12} \\
  Improvement& 2.18$\times$ & 1.29$\times$ & 0.95 $(\downarrow)$ \\
\midrule
GraphSAINT-RW & 108.9 & 0.65 & 51.03 \scriptsize{$\pm$ 0.20} \\
\textbf{\sg}         & 37.3  & 0.49 & 50.15 \scriptsize{$\pm$ 0.24} \\
  Improvement&2.92$\times$ & 1.33$\times$ & 0.88 $(\downarrow)$\\
\bottomrule
\end{tabular}
\end{small}
\end{center}
\end{table}

\begin{figure*}[!t]
  \centering
  \begin{minipage}[b]{0.24\textwidth}
    \includegraphics[width=\textwidth]{figs/arxiv.pdf}
  \end{minipage}
  \begin{minipage}[b]{0.24\textwidth}
    \includegraphics[width=\textwidth]{figs/reddit.pdf}
  \end{minipage}
  \begin{minipage}[b]{0.24\textwidth}
    \includegraphics[width=\textwidth]{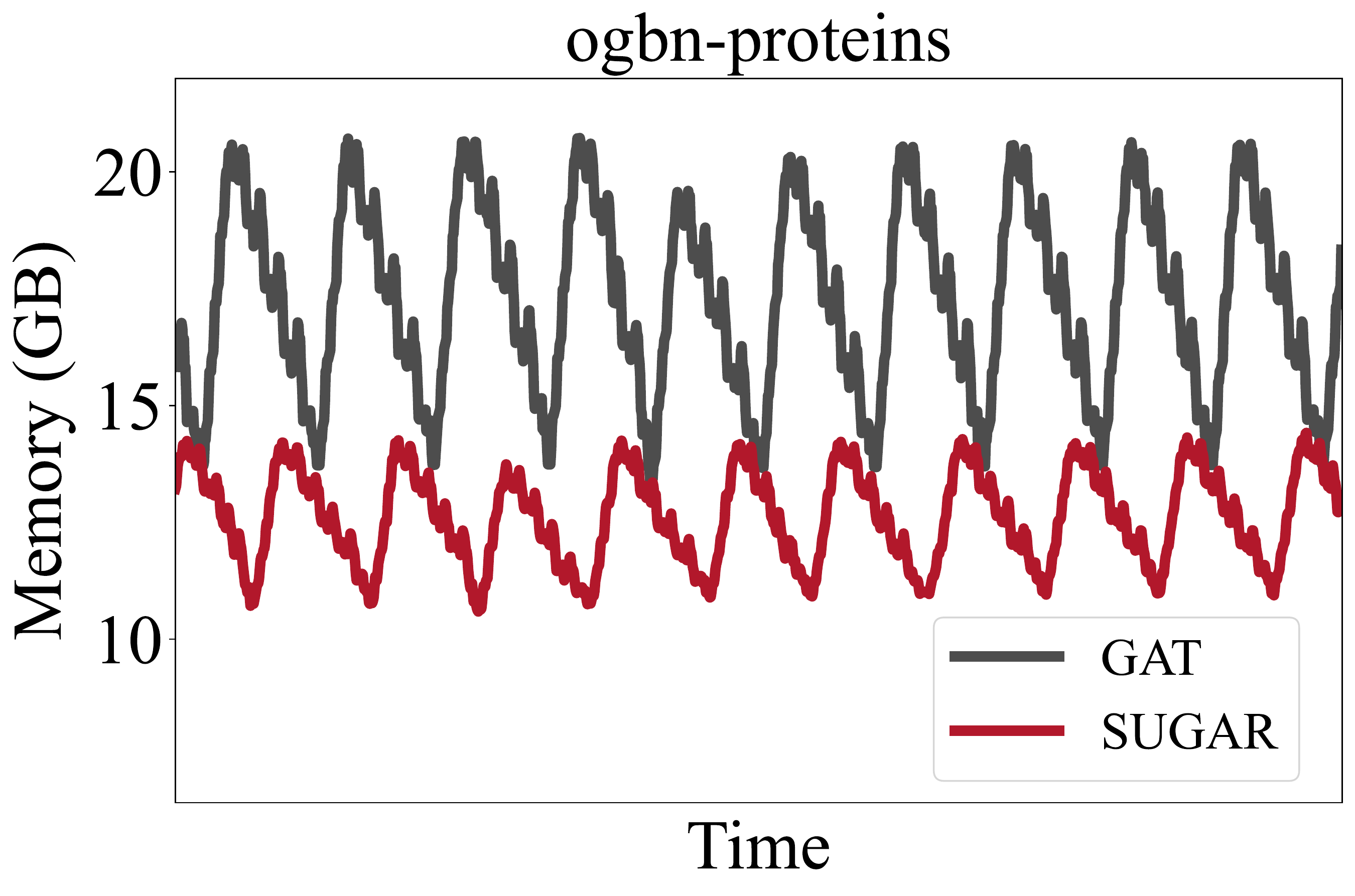}
  \end{minipage}
  \begin{minipage}[b]{0.24\textwidth}
    \includegraphics[width=\textwidth]{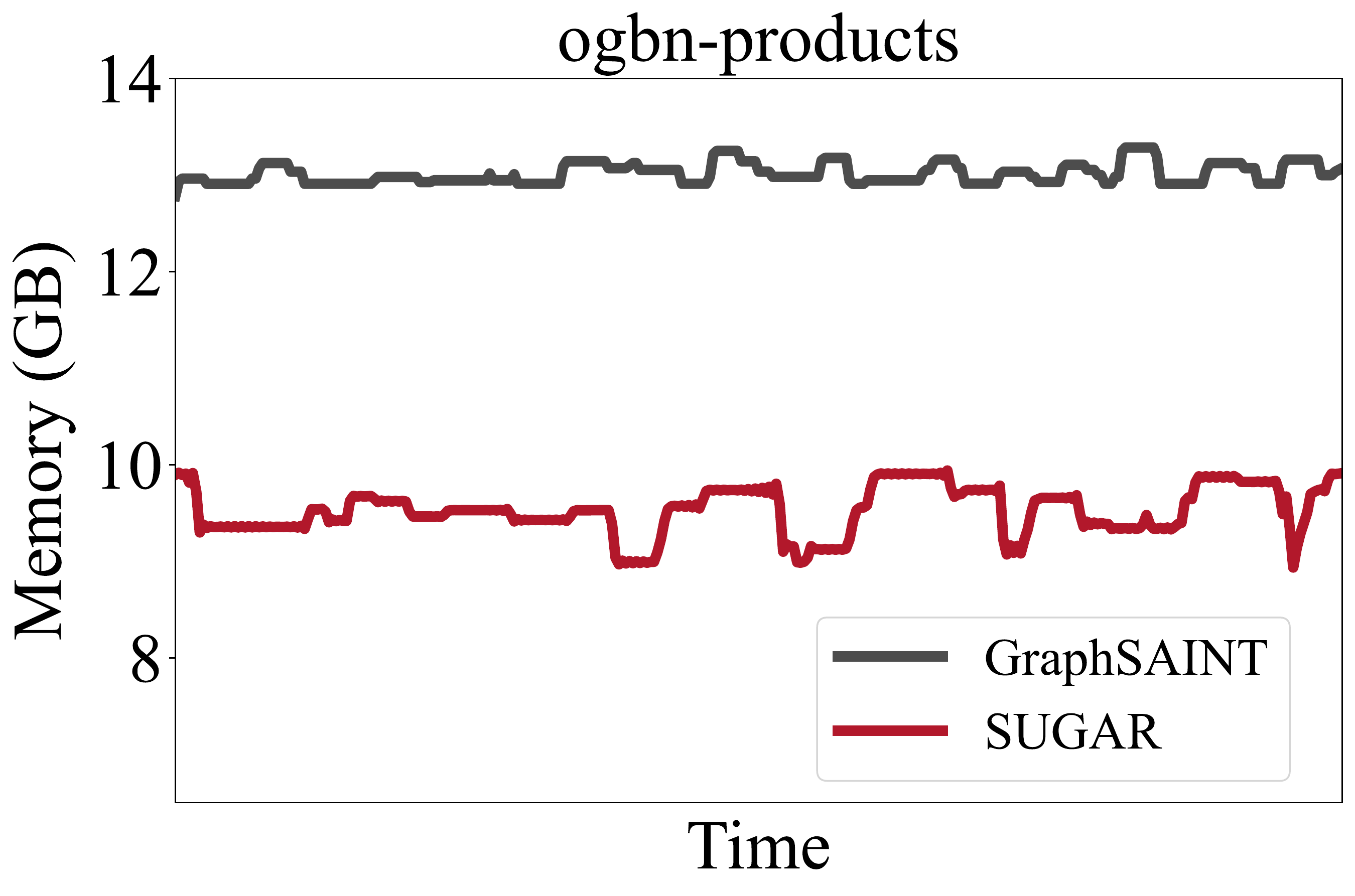}
  \end{minipage}
  \caption{Memory variation during training GNNs on Desktop-CPU for the four datasets: \textit{ogbn-arxiv}, \textit{Reddit}, \textit{ogbn-proteins} and \textit{ogbn-products}. For \textsc{SUGAR}, we plot the memory variation of the device that consumes most memory. As it can be seen, the memory requirements for \sg are significantly smaller than baseline methods.}.
  \label{fig.mem2}
\end{figure*}

Figure \ref{fig.mem2} compares the resident set size (RSS) memory usage of \sg against GNN baselines on a CPU setting on the four datasets: \textit{Reddit}, \textit{ogbn-arxiv}, \textit{ogbn-proteins} and \textit{ogbn-products}. We train a full-batch version of GCN and the batch size of GAT is larger compared with GraphSAGE and GraphSAINT. This accounts for higher fluctuation in the corresponding figure. It is evident that our proposed \sg achieves substantial memory reductions compared with baseline GNNs.

We present a case study of \sg on NVIDIA Jetson Nano in Table \ref{tab.jetson}. Jetson Nano is a popular, cheap and readily available platform (we adopt the model with quad Cortex-A57 CPU and 4GB LPDDR memory) and thus considered as a good fit for our problem scenario. Apart from training time, we measure the peak RSS memory usage for the training process and calculate energy consumption. As shown in Table \ref{tab.jetson}, \sg achieves low latency, consumes less memory and is more energy efficient than baseline GNN algorithms. Therefore, it provides an ideal choice to train GNNs on devices with limited memory and battery capacity.

\begin{table}[!tbh]
\caption{Evaluations of \sg on NVIDIA Jetson Nano for \textit{Flickr} and \textit{ogbn-arxiv}. `Avg. Time' and `Max Mem' denotes training time per epoch and peak resident set size (RSS) memory. We measure the time, memory and energy for training 10 epochs. \sg improves average training time, memory usage and energy consumption per device over baseline GNNs (\ie, GraphSAINT and GCN).} 
\label{tab.jetson}
\vskip 0.1in
\begin{center}
\begin{small}
\begin{tabular}{l|lccc}
\toprule
Dataset & &  \makecell[c]{Avg. Time \\ \lbrack sec\rbrack} & \makecell[c]{Max Mem \\ \lbrack GB\rbrack} & \makecell[c]{Energy \\ \lbrack kJ\rbrack}\\
 \midrule
\multirow{2}{*}{\textit{Flickr}} & GraphSAINT-N  & 22.62 & 1.05 & 1.13 \\
~ & \textbf{\sg} & 10.50 & 0.89 & 0.52 \\
~ & Improvement & 2.15$\times$ & 1.18$\times$ & 2.17$\times$ \\
\midrule
\midrule
\textit{ogbn-} & GCN & 28.10 & 2.24 & 1.27 \\
\textit{arxiv} & \textbf{\sg} & 18.39 & 1.46 & 0.81 \\
~ & Improvement & 1.53$\times$ & 1.53$\times$ & 1.57$\times$\\
\bottomrule
\end{tabular}
\end{small}
\end{center}
\vskip -0.15in
\end{table}

\section{Scalability Analysis (Section \ref{sec.exp}, Page 7 in the paper)}\label{apsec.scale}
\subsection{Number of partitions}
\begin{figure*}[!tb]
  \centering
    \begin{minipage}[b]{0.32\textwidth}
      \includegraphics[width=\textwidth]{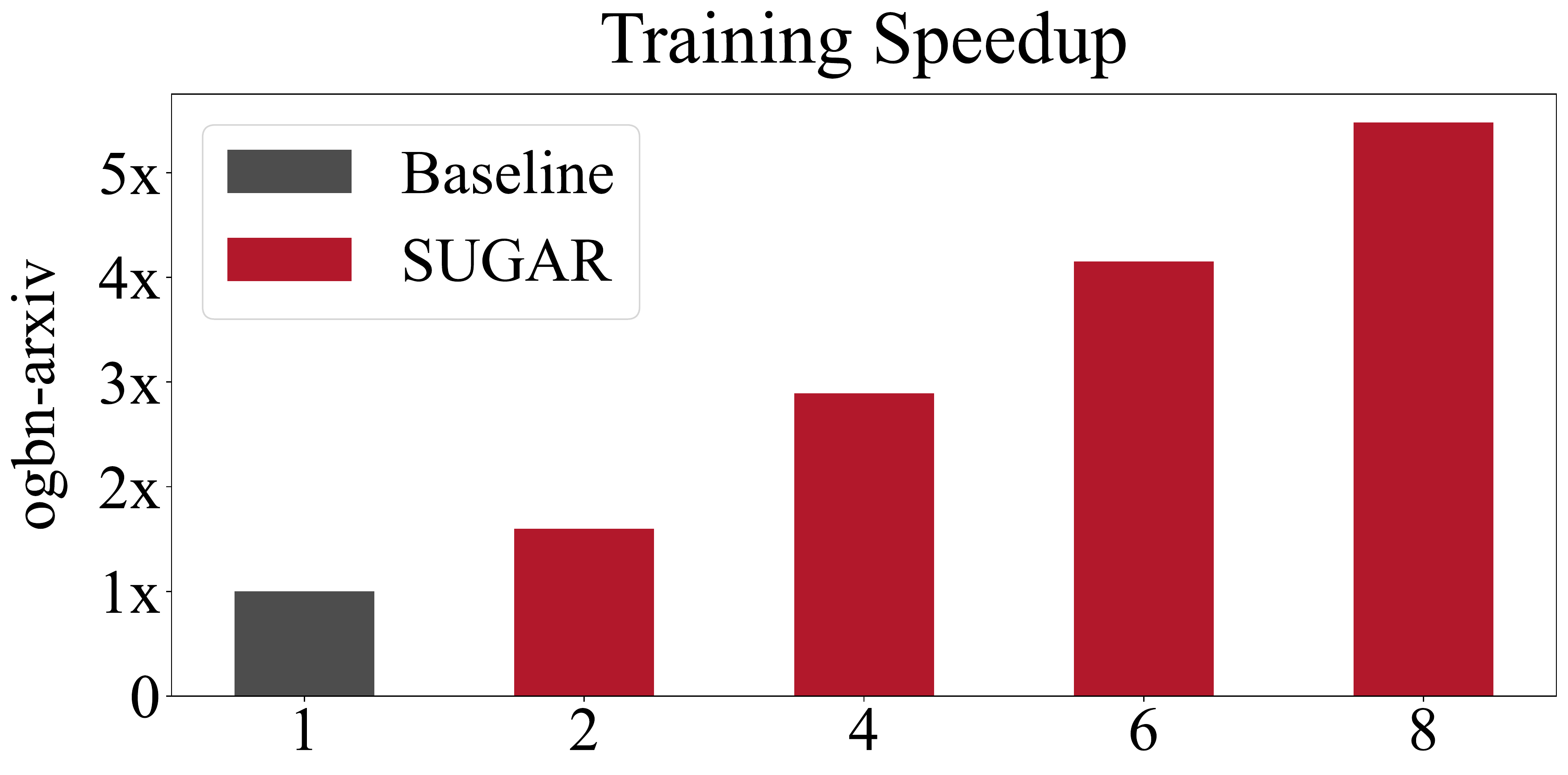}
    \end{minipage}
    \hspace{0.4mm}
    \begin{minipage}[b]{0.32\textwidth}
      \includegraphics[width=\textwidth]{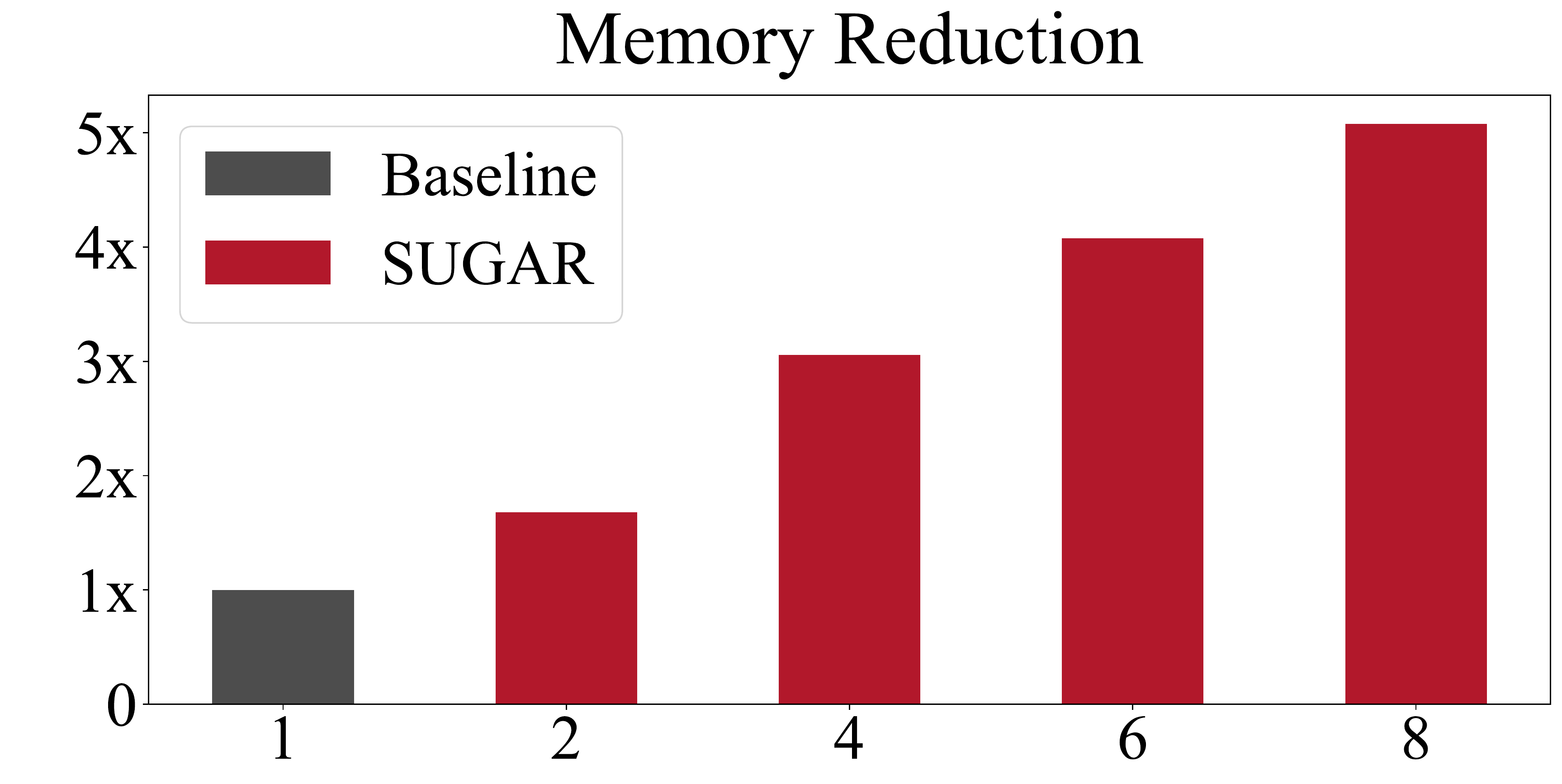}
    \end{minipage}
    \begin{minipage}[b]{0.32\textwidth}
      \includegraphics[width=\textwidth]{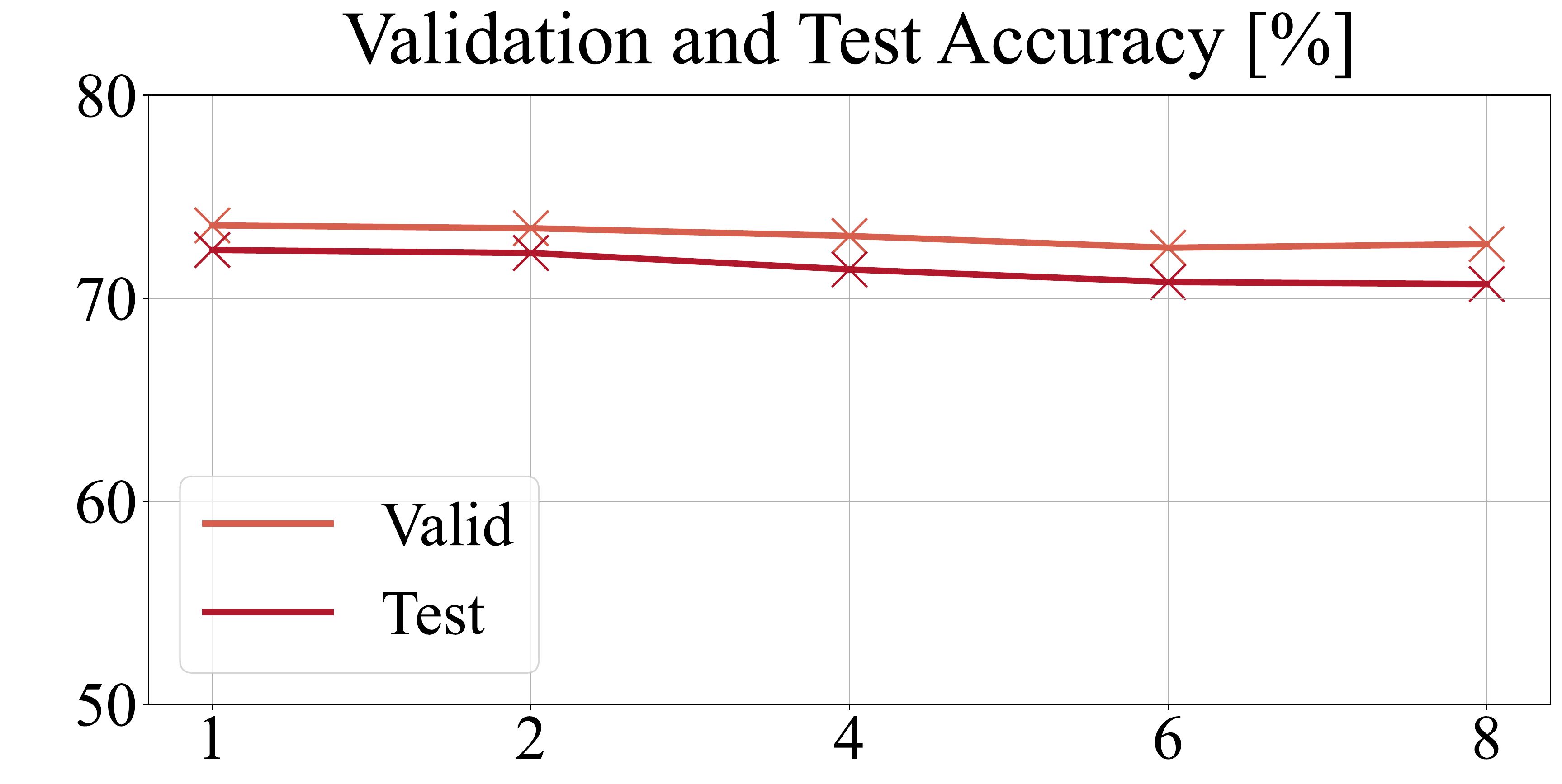}
    \end{minipage}
    
    \begin{minipage}[b]{0.32\textwidth}
      \includegraphics[width=\textwidth]{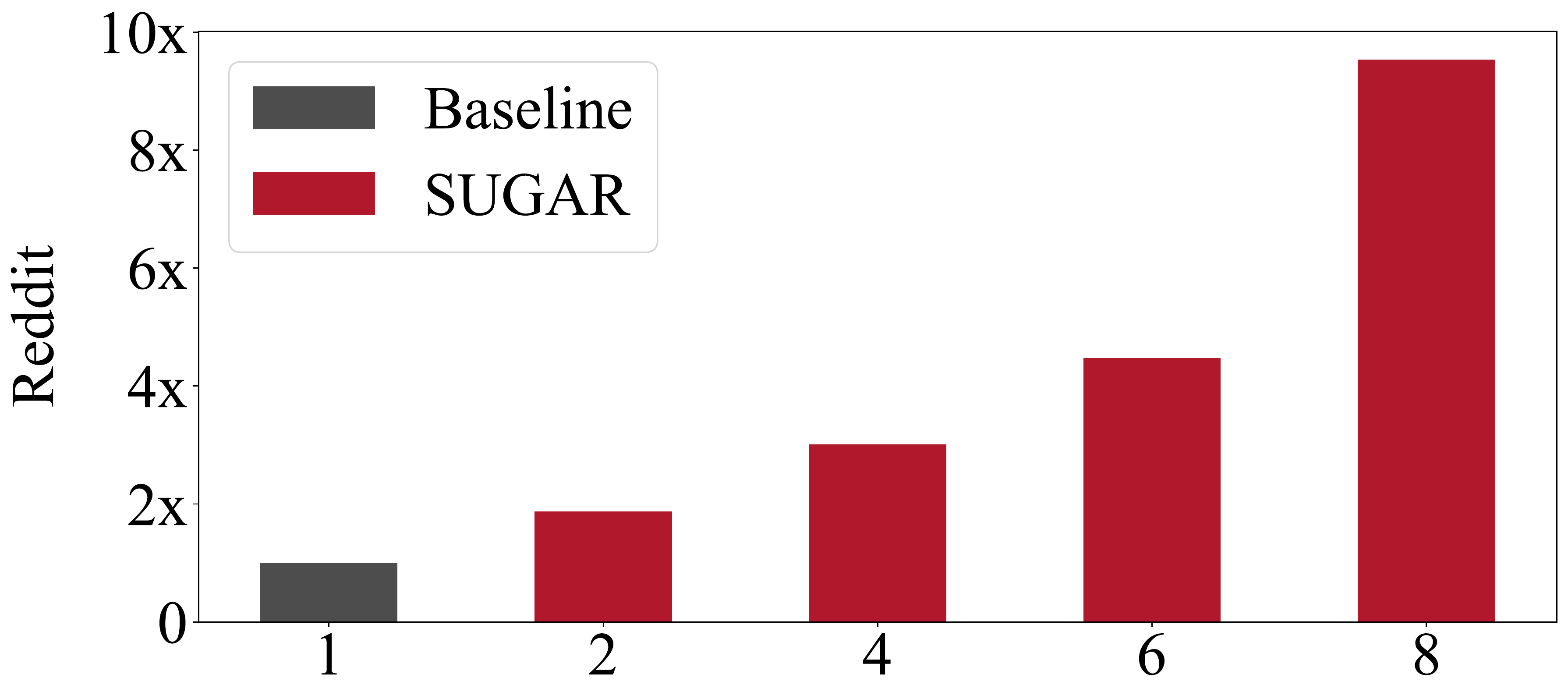}
    \end{minipage}
    \begin{minipage}[b]{0.32\textwidth}
      \includegraphics[width=\textwidth]{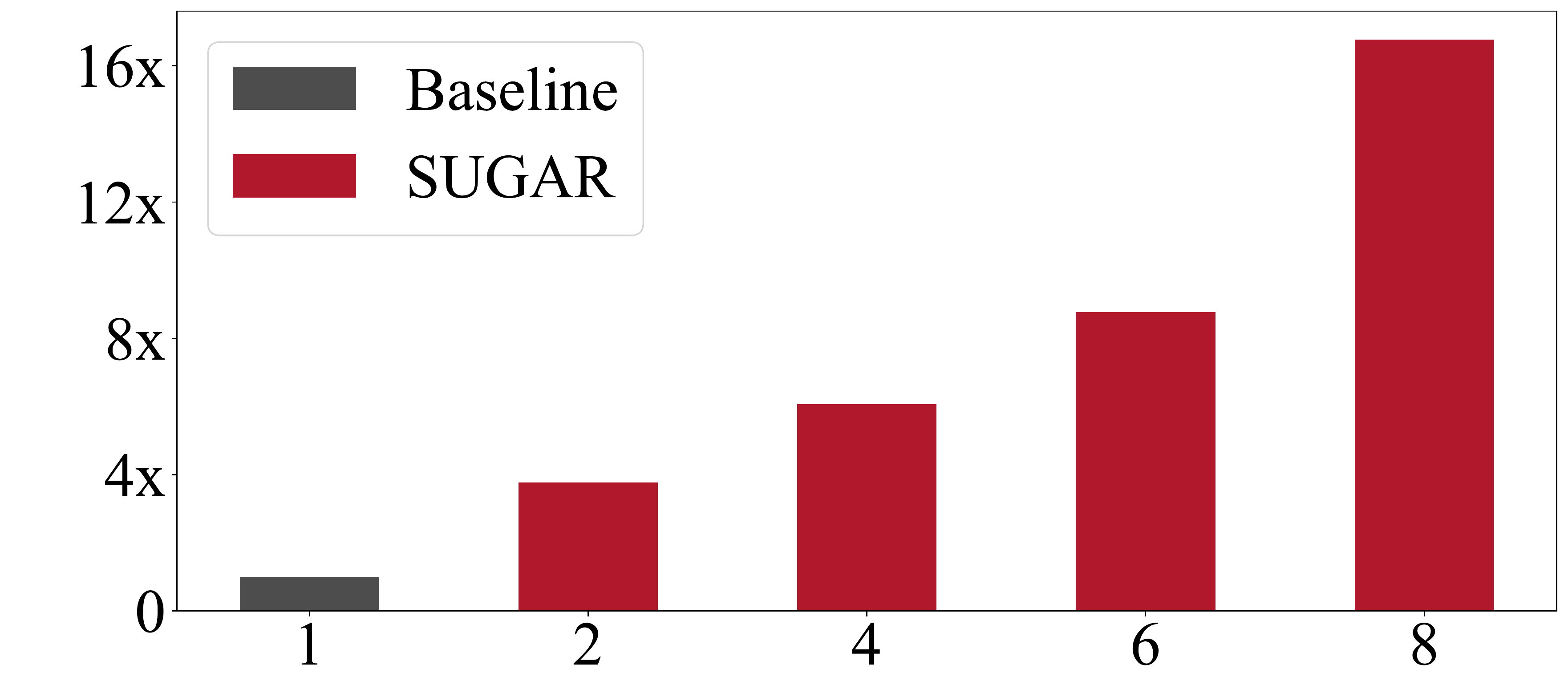}
    \end{minipage}
    \begin{minipage}[b]{0.32\textwidth}
      \includegraphics[width=\textwidth]{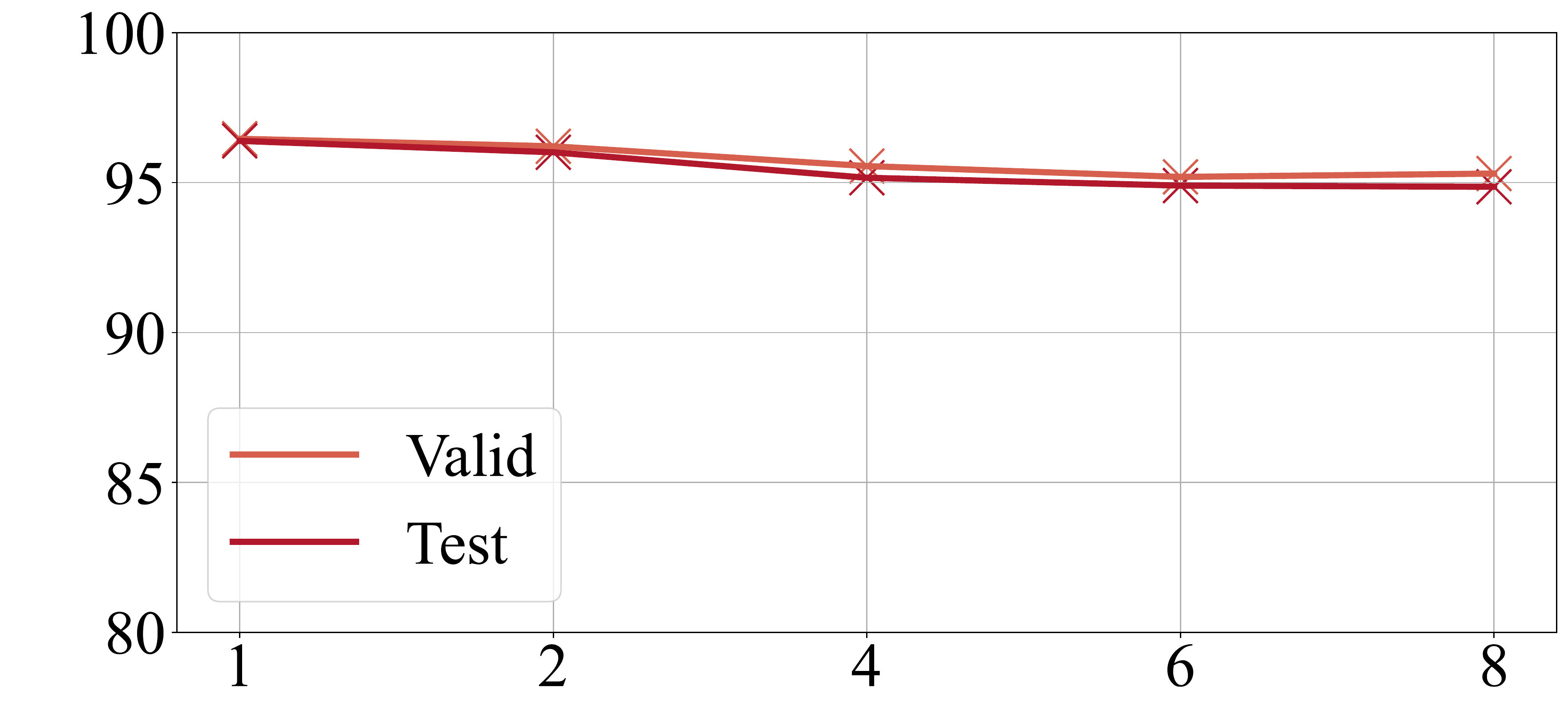}
    \end{minipage}
    
    \begin{minipage}[b]{0.32\textwidth}
      \includegraphics[width=\textwidth]{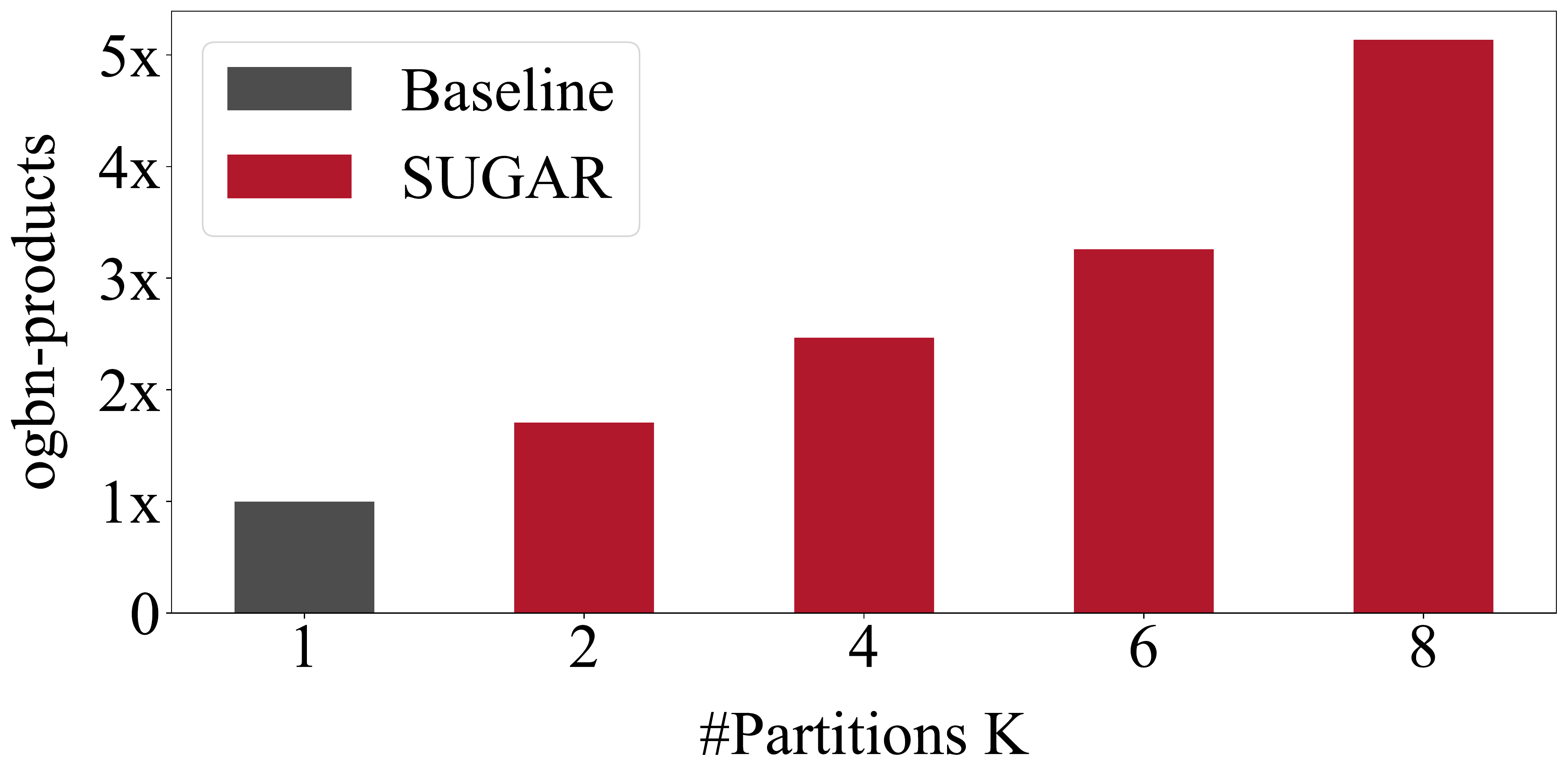}
    \end{minipage}
    \hspace{0.4mm}
    \begin{minipage}[b]{0.32\textwidth}
      \includegraphics[width=\textwidth]{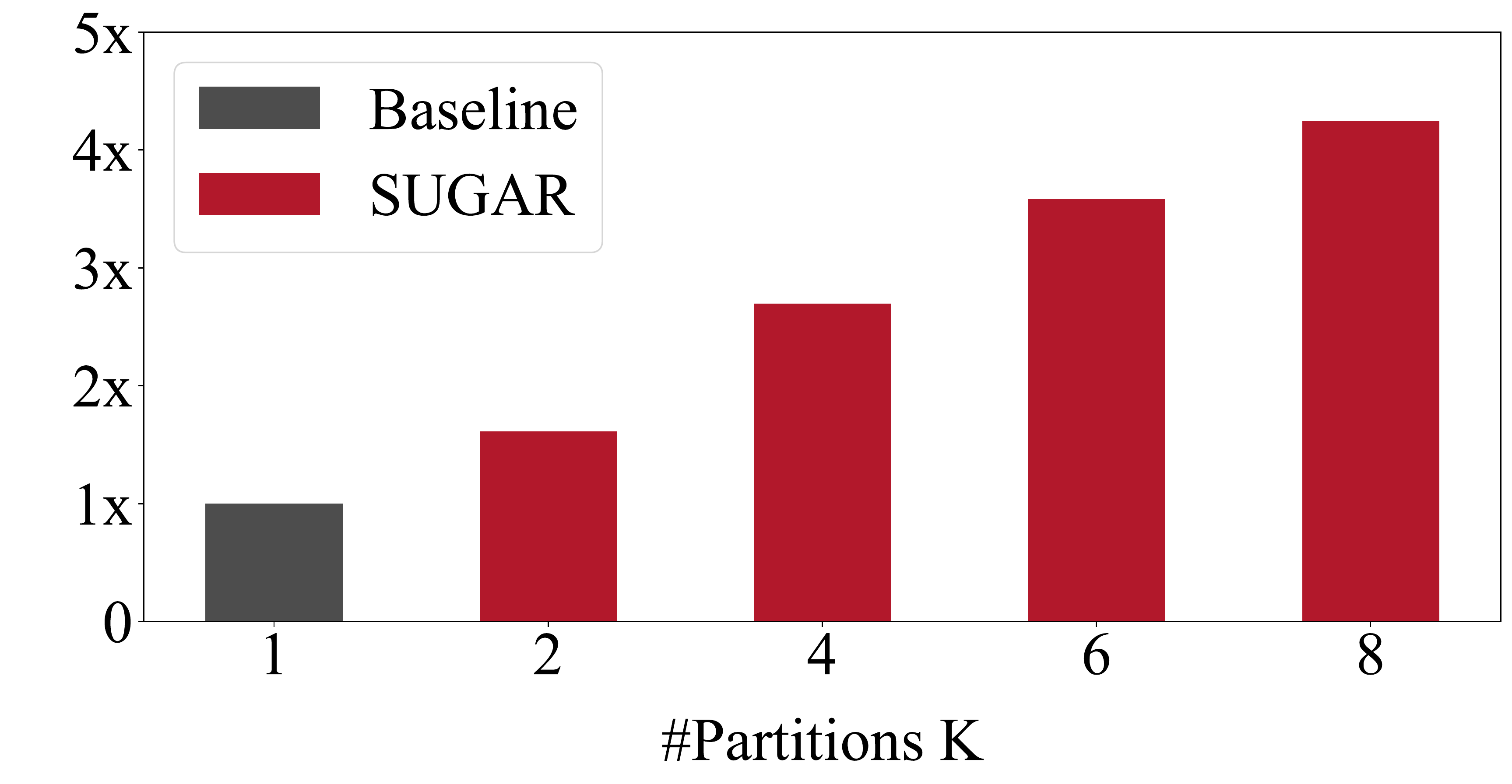}
    \end{minipage}
    \begin{minipage}[b]{0.32\textwidth}
      \includegraphics[width=\textwidth]{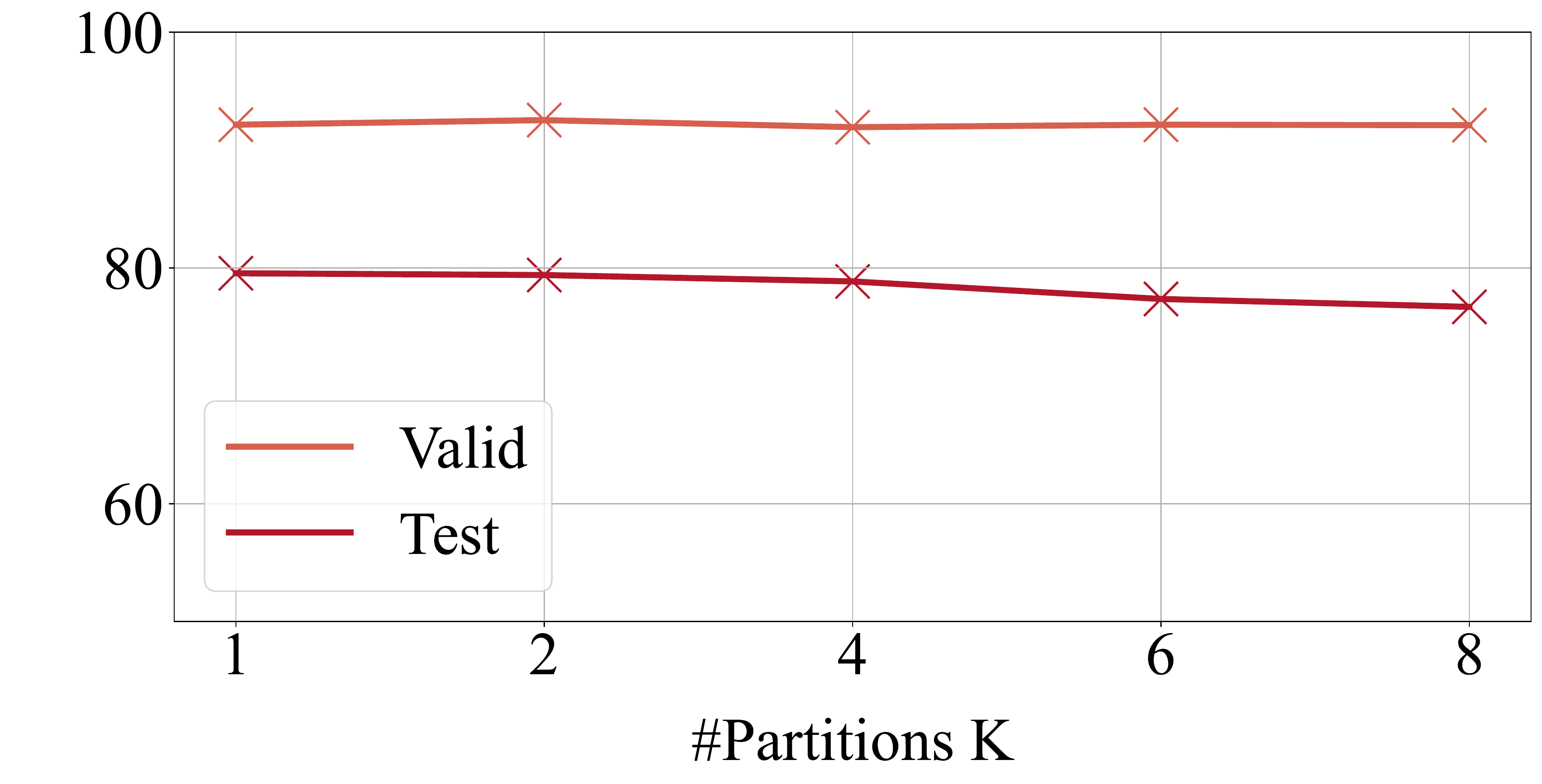}
    \end{minipage}
  \caption{Scalability analysis on the number of partitions (\ie, the number of available devices $K$) for \textsc{SUGAR}. $K=1$ refers to the baseline GNN (\ie, GCN for \textit{ogbn-arxiv}; GraphSAGE for \textit{Reddit}; GraphSAINT for \textit{ogbn-products}). We report the smallest training time speedup and peak GPU memory reduction among $K$ devices (\ie, the worst-case scenario) of \sg over the baseline.}
  \label{fig.abl1}
\end{figure*}

Below we provide a scalability analysis of \sg based on the number of partitions (\ie, device number $K$).

We vary the number of available devices $K$ from 2 to 8 and evaluate \sg on the \textit{ogbn-arxiv}, \textit{Reddit} and \textit{ogbn-products} datasets. The evaluation is conducted on Desktop-GPU. Runtime speedup, peak GPU memory reduction, validation and test accuracy are presented in Figure \ref{fig.abl1}. With increasing $K$, we observe a decreased training time and peak memory usage for each local device. 

As we can see, while distributing the GNN model to more devices yields computation efficiency, test accuracy drops a bit. For instance, in the case of 8 devices, the biggest decrease happens in the \textit{ogbn-products} dataset: test accuracy is 76.69\% while the baseline accuracy is 79.54\%. In the meantime, \sg leads to $5.13\times$ speedup, as well as $4.24\times$ memory reduction compared with the baseline. Generally speaking, there exists a tradeoff between training scalability and performance. The underlying reason is that the increase of partition number $K$ leads to more inter-device edges, which corresponds to a larger error in estimating with $A_{SG}$ with $A$. 

We further evaluated \sg in a 128-device setting. The results show that the test accuracy drop compared with baseline GNNs is small, \ie, within 5\% when scaling up to 128 devices (\eg, accuracy decreases from 72.37\% to 67.80\% for \textit{ogbn-arxiv}, from 96.39\% to 92.32\% for \textit{Reddit}, from 50.64\% to 46.31\% for \textit{Flickr}). At the same time, we note that the memory savings are great (\eg, peak memory usage per device is reduced from 1.60GB to 0.02GB for \textit{ogbn-arxiv}). This shows that \sg can work with very small computation and memory requirements at the cost of slightly downgraded performance. Thus, this provides a feasible solution in extremely resource-limited scenarios while general GNN training methods are not applicable.

\subsection{Batch Size}
\begin{figure*}[!t]
  \centering
  \subfigure[graph data on CPU]{
    \begin{minipage}[b]{0.25\textwidth}
      \includegraphics[width=\textwidth]{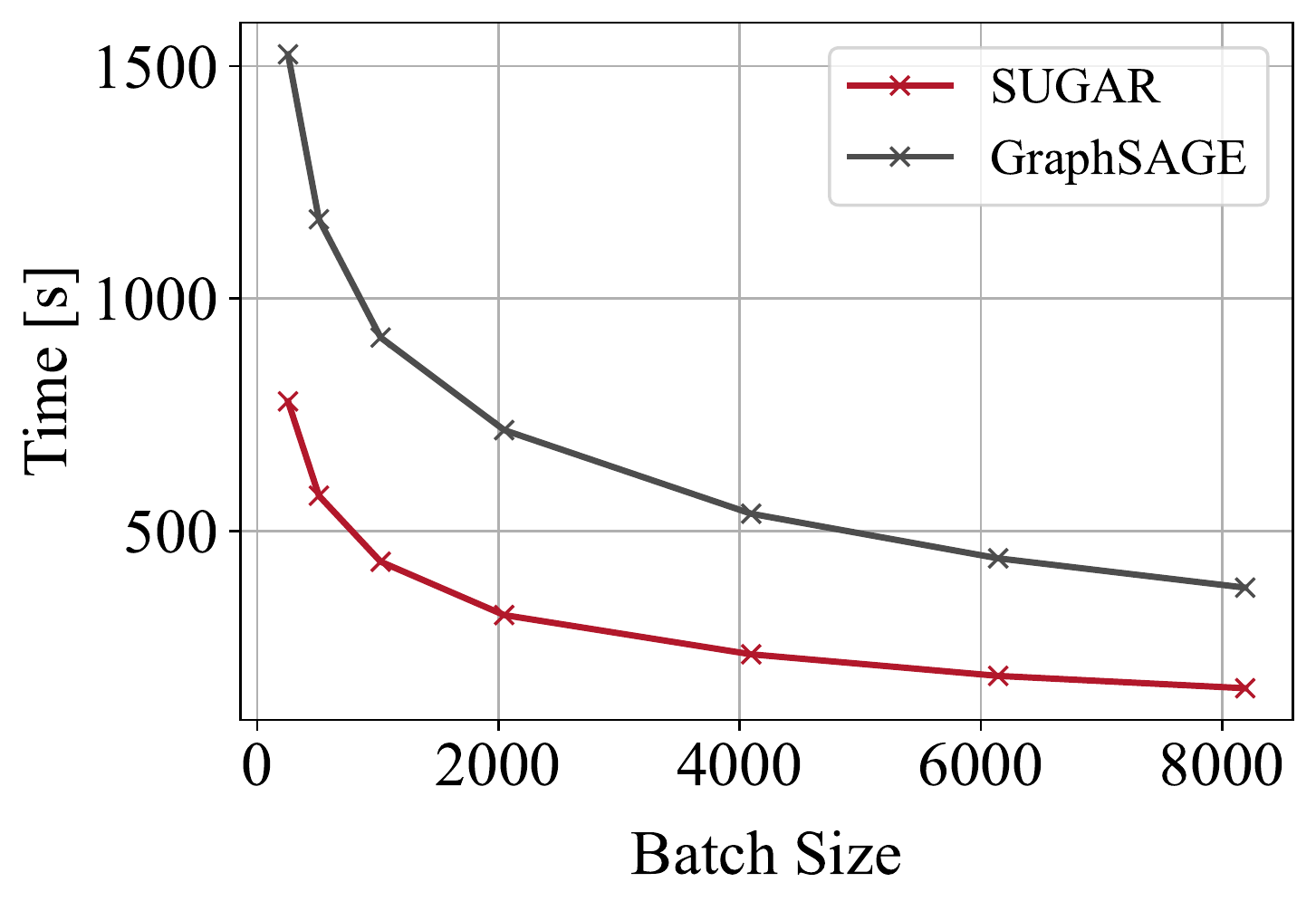}
    \end{minipage}
    \begin{minipage}[b]{0.23\textwidth}
      \includegraphics[width=\textwidth]{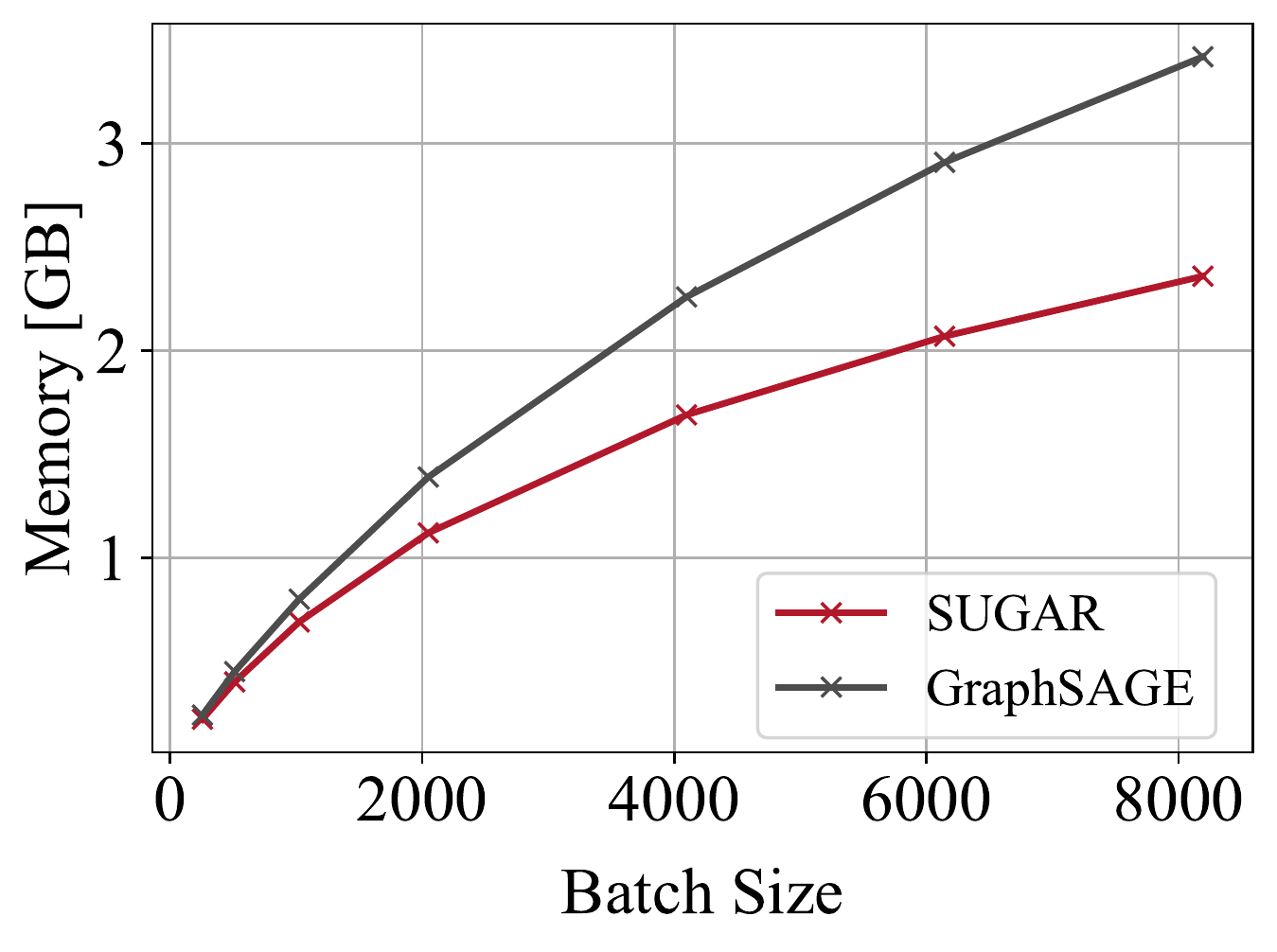}
    \end{minipage}
  }
  \subfigure[graph data on GPU]{
    \begin{minipage}[b]{0.24\textwidth}
      \includegraphics[width=\textwidth]{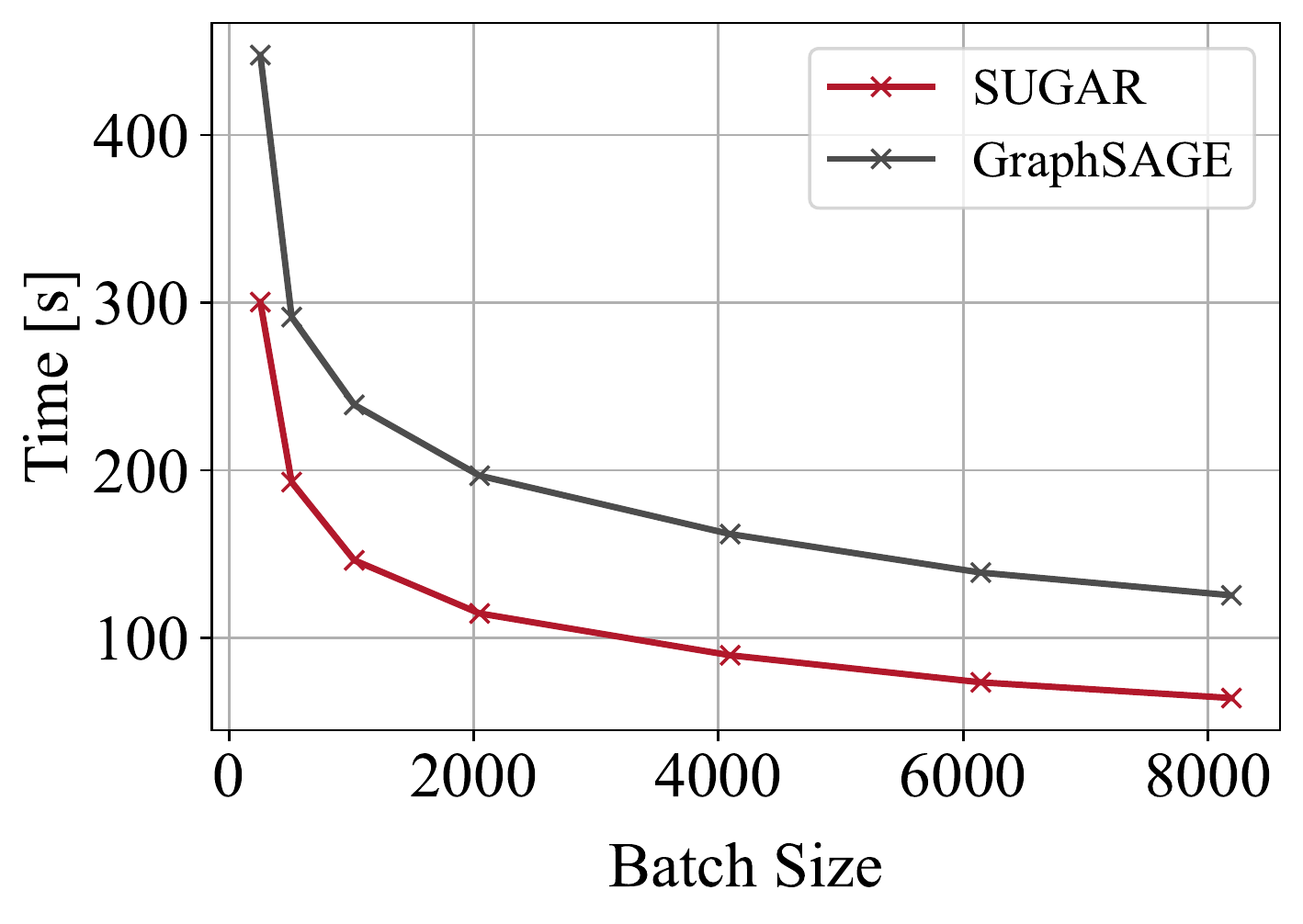}
    \end{minipage}
    \begin{minipage}[b]{0.23\textwidth}
      \includegraphics[width=\textwidth]{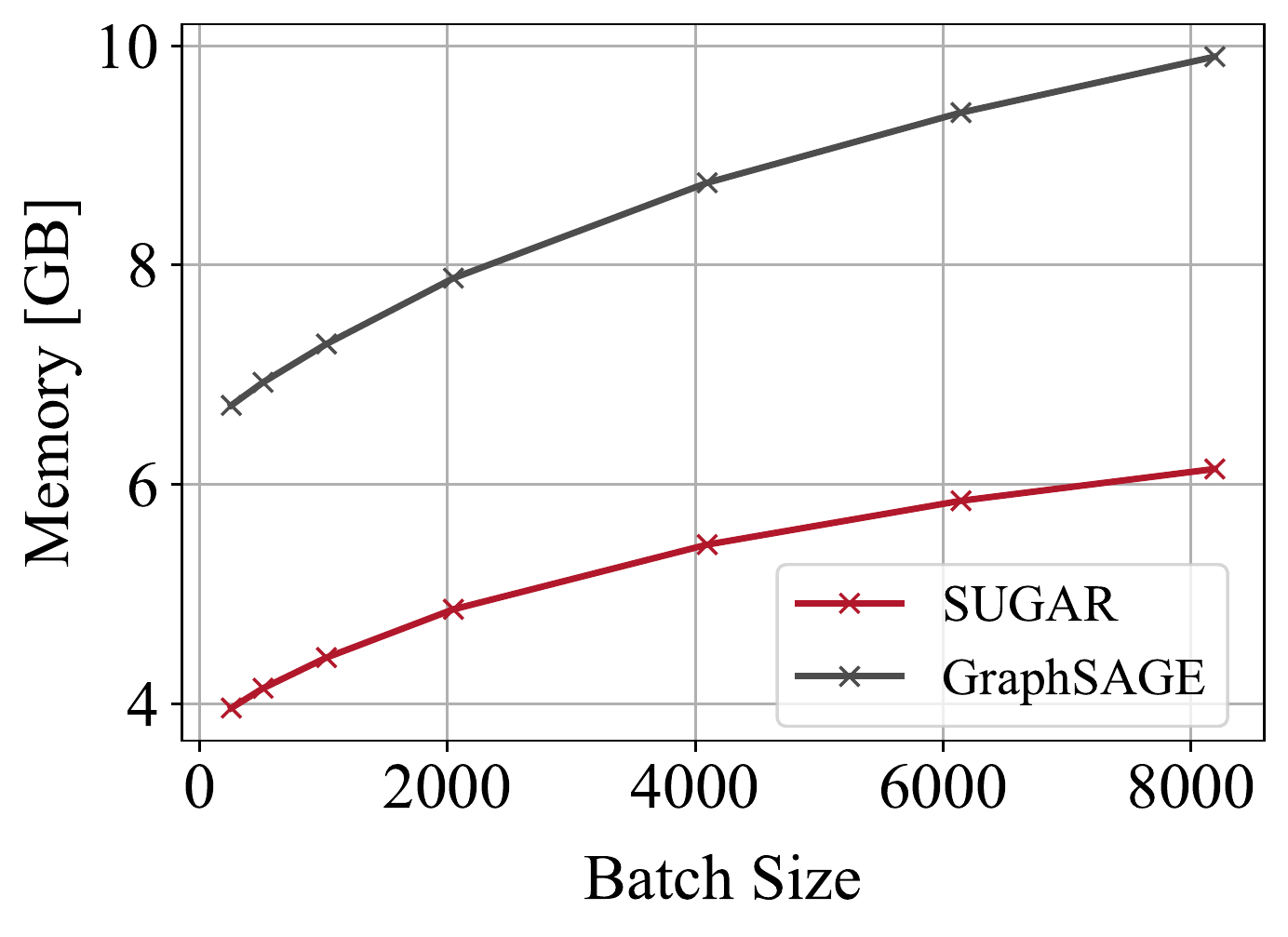}
    \end{minipage}
  }
  \caption{The training time and peak GPU memory with varying batch sizes of GraphSAGE and \sg for the \textit{ogbn-products} data. We investigate two settings: (a) graph data loaded on GPU for faster execution (b) graph data loaded on CPU for memory savings.}
  \label{fig.abl2}
\end{figure*}
For mini-batch training algorithms, when the limited memory of device renders GNN training infeasible, a natural idea is reduce the batch size for memory savings. Here, we analyze the influence of \sg and the act of reducing batch sizes on computational efficiency, as well as memory scalability. We conduct experiments on the largest \textit{ogbn-products} graph with GraphSAGE as the baseline. Two settings are considered: (a) graph data loaded on CPU, longer training time and smaller memory consumption is expected; (b) graph data loaded on GPU, the model runs faster, yet requires more GPU memory. Figure \ref{fig.abl2} in this Appendix provides runtime and memory results with varying batch sizes.

We have two observations: (1) \sg mainly improves runtime in setting (a) and achieves greater memory reduction in setting (b). This is related to the mechanism of \textsc{SUGAR}: each local model adopts one subgraph for training instead of the original graph; thus, data loading time is reduced in setting (a) and putting a subgraph on GPU is more memory efficient in setting (b). (2) \sg demonstrates to be a better technique in reducing memory usage than tuning the batch size. While it is generally known that there exists a tradeoff between computation and memory requirements as reducing batch size increases training time, \sg is able to improve on both accounts.
